\documentclass{article}

\usepackage{arxiv}

\usepackage[utf8]{inputenc} 
\usepackage[T1]{fontenc}    
\usepackage{hyperref}       
\usepackage{url}            
\usepackage{booktabs}       
\usepackage{amsfonts}       
\usepackage{nicefrac}       
\usepackage{microtype}      
\usepackage{lipsum}		
\usepackage{graphicx}
\usepackage{natbib}
\usepackage{doi}
\usepackage{subcaption}
\usepackage{setspace}

\usepackage{bm}
\usepackage{algorithm}
\usepackage{algpseudocode}
\usepackage{mathtools} %
\usepackage{dsfont} %
\usepackage{bbm}
\usepackage{booktabs} %
\usepackage{diagbox} 
\usepackage{nccmath} %
\usepackage{arydshln} %
\usepackage{tablefootnote}
\usepackage{bbding} %
\usepackage{amssymb}
\usepackage{enumitem} %
\usepackage{amsmath}

\usepackage{xcolor}
\usepackage{caption}
\usepackage{subcaption}
\usepackage{wrapfig}
\usepackage{multirow}
\usepackage{soul}

\usepackage{stackengine}

\newcommand{\TE}{T\!E}

\newcommand{\NDE}{N\!D\!E}

\newcommand{\TIE}{T\!I\!E}
\newcommand{\softmax}{\text{softmax}}

\renewcommand{\caption}[1]{\vspace{-3mm}\caption{#1}\vspace{-3mm}}

\title{Unveiling Cross Modality Bias in Visual Question Answering: A Causal View with Possible Worlds VQA}

\author{
    Ali Vosoughi\textsuperscript{3}\thanks{Equal contribution}, 
    Shijian Deng\textsuperscript{2}\footnotemark[1], 
    Songyang Zhang\textsuperscript{1}, 
    Yapeng Tian\textsuperscript{2}, 
    Chenliang Xu\textsuperscript{1},  
    Jiebo Luo\textsuperscript{1,3} \\
    \vspace{0.5em}
    \textsuperscript{1}Dept. of Comp. Sci., University of Rochester, Rochester, NY 14620 \\
    \textsuperscript{2}Dept. of Comp. Sci., University of Texas Dallas, Dallas, TX 12345 \\
    \textsuperscript{3}Dept. of ECE, University of Rochester, Rochester, NY 14620 
}

\renewcommand*\footnoterule{
  \kern-3pt
  \hrule width 0.4\columnwidth
  \kern2.6pt
}

\date{}

\renewcommand{\shorttitle}{PWVQA: Possible Worlds VQA}

\hypersetup{
pdftitle={Unveiling Cross Modality Bias in Visual Question Answering: A Causal View with Possible Worlds VQA},
pdfsubject={q-bio.NC, q-bio.QM},
pdfauthor={David S.~Hippocampus, Elias D.~Striatum},
pdfkeywords={First keyword, Second keyword, More},
}

\begin{document}

\maketitle
\begin{center}
    \centering
    \captionsetup{type=figure}
    \includegraphics[width=.325\linewidth]{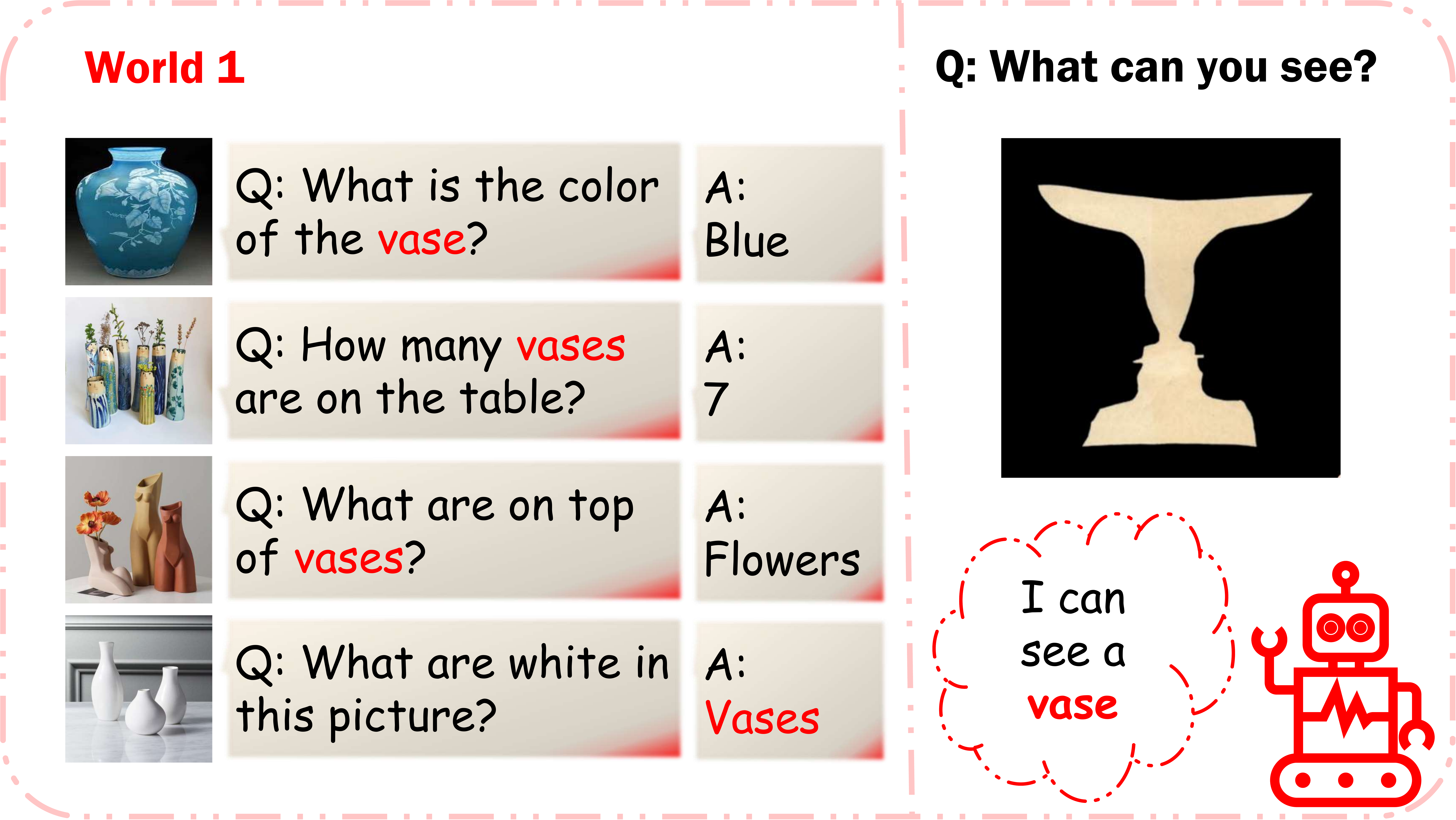}
    \includegraphics[width=.325\linewidth]{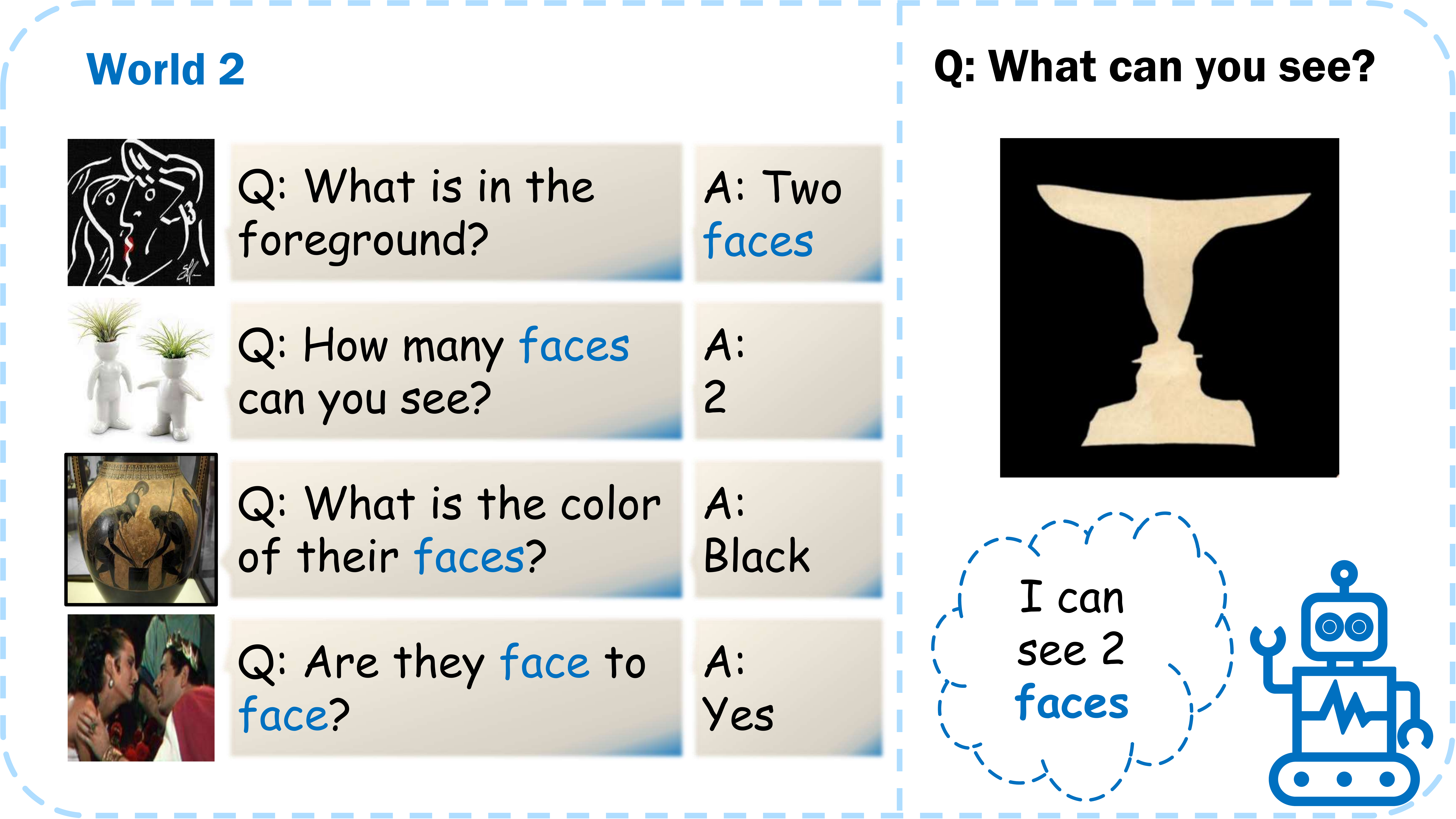}
    \includegraphics[width=.325\linewidth]{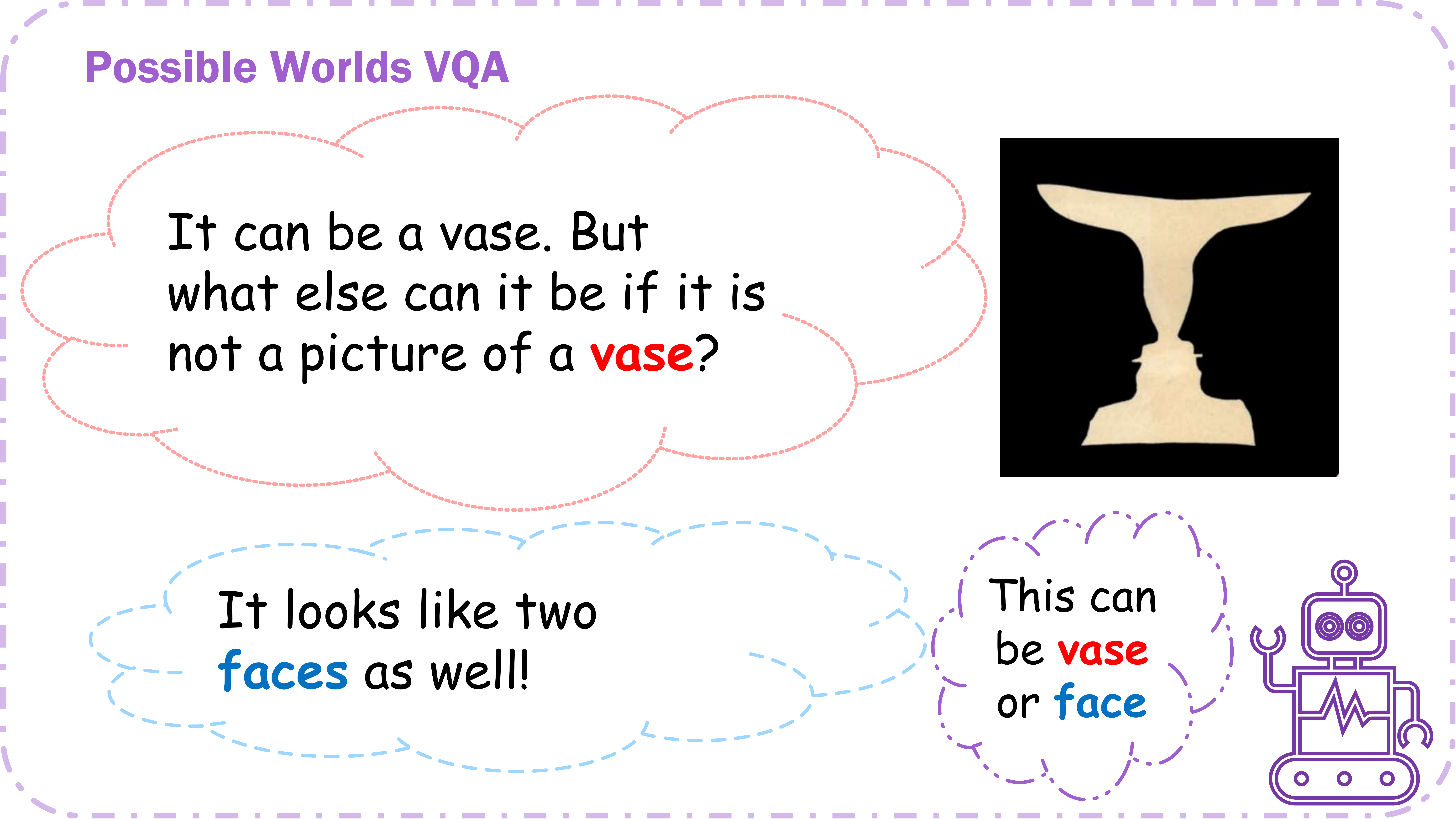}
    \captionof{figure}{Visual cognition of annotators is related to the language~\cite{lupyan2020effects}, which in turn influences the choice of questions and answers to those questions. A well-known example of Rubin's vase shown in this figure~\cite{rubin1915synsoplevede} is to illustrate how memory and experience can affect the perception of the annotator.}\label{fig:teaser_pwvqa}
    
\end{center}%

\begin{abstract}
	To increase the generalization capability of VQA systems, many recent studies have tried to de-bias spurious language or vision associations that shortcut the question or image to the answer. 
Despite these efforts, the literature fails to address the confounding effect of vision and language simultaneously. 
As a result, when they reduce bias learned from one modality, they usually increase bias from the other.
In this paper, we first model a confounding effect that causes language and vision bias simultaneously, then propose a counterfactual inference to remove the influence of this effect. 
The model trained in this strategy can concurrently and efficiently reduce vision and language bias. 
To the best of our knowledge, this is the first work to reduce biases resulting from confounding effects of vision and language in VQA, leveraging causal explain-away relations.
We accompany our method with an explain-away strategy, pushing the accuracy of the questions with numerical answers results compared to existing methods that have been an open problem. 
The proposed method outperforms the state-of-the-art methods in VQA-CP v2 datasets.

\end{abstract}


\section{Introduction}\label{sec:introduction}
Visual Question Answering (VQA) systems are one of the most fundamental building blocks at the intersection of vision and language~\cite{zellers2019recognition, niu2021counterfactual,kolling2022efficient}. 
VQA systems use linguistic and visual information to obtain correct and robust answers to given questions from an image. 
Despite the efforts, regrettably, most VQA systems shortcut directly from the vision or language to an answer~\cite{ niu2021introspective, cadene2019rubi, cadene2019murel}.
This shortcut is known as vision or language bias and has been well-studied in recent years~\cite{jing2020overcoming, ramakrishnan2018overcoming, cadene2019rubi, clark2019don, niu2021counterfactual, gat2020removing}. 

Spurious correlations sometimes shortcut an answer to an image, and other times question to an answer. CF-VQA~\cite{niu2021counterfactual} was proposed to alleviate this problem by replacing natural indirect effect (NIE) with total indirect effect (TIE). Still, this method focuses only on language bias, ignores the visual information, and can also mislead the VQA model, resulting in CF-VQA sometimes giving its answer directly by uttering salient objects in the picture even if the answer obviously should be a number or "yes/no" according to the type of the question. 

Recent studies suggest that memory and culture influence the perception of visual information in humans~\cite{lupyan2020effects}, the problem that we illustrate in Fig.~\ref{fig:teaser_pwvqa} with a famous example of Rubin's vase~\cite{rubin1915synsoplevede}, where the same image can be perceived differently. The difference in preference and perception confounds VQA datasets, making them biased in data collection and annotation process~\cite{VQA, niu2021counterfactual}. Therefore, VQA models fail to generalize as these confounders affect vision and language in datasets.

Contradicting those existing methods, we propose a new system called possible worlds VQA (PW-VQA) to address vision and language biases by removing the confounding effects of two modalities through a causal lens. After removing these effects captured through training, our model is less biased by either language or vision modality during test time. Furthermore, compared to other models, ours achieved significant performance improvement on the numerical questions, which used to be a struggling problem for previous methods.

Our contributions are as follows. 1) We propose a causal graph separating the problem into two sub-graphs of anticausal learning and an explain-away network. We simultaneously model the visual and linguistic biases through the explain-away network to distinguish between bad and good language and vision biases. We model the experience bias of the annotator as an unobserved confounder that influences the choice of question and answer pairs. 2) We propose a counterfactual approach to reduce these bad biases while keeping the good ones. To the best of our knowledge, our work is the first to propose a causal method to simultaneously alleviate language and vision biases. 3) We double the accuracy of the numerical questions, which has been an open question recently~\cite{niu2021counterfactual}.

\section{Motivation and Background}\label{sec:relatedwork}

\begin{figure*}[t]
     \centering
     \begin{subfigure}[t]{0.22\textwidth}
         \centering
         \includegraphics[width=\textwidth]{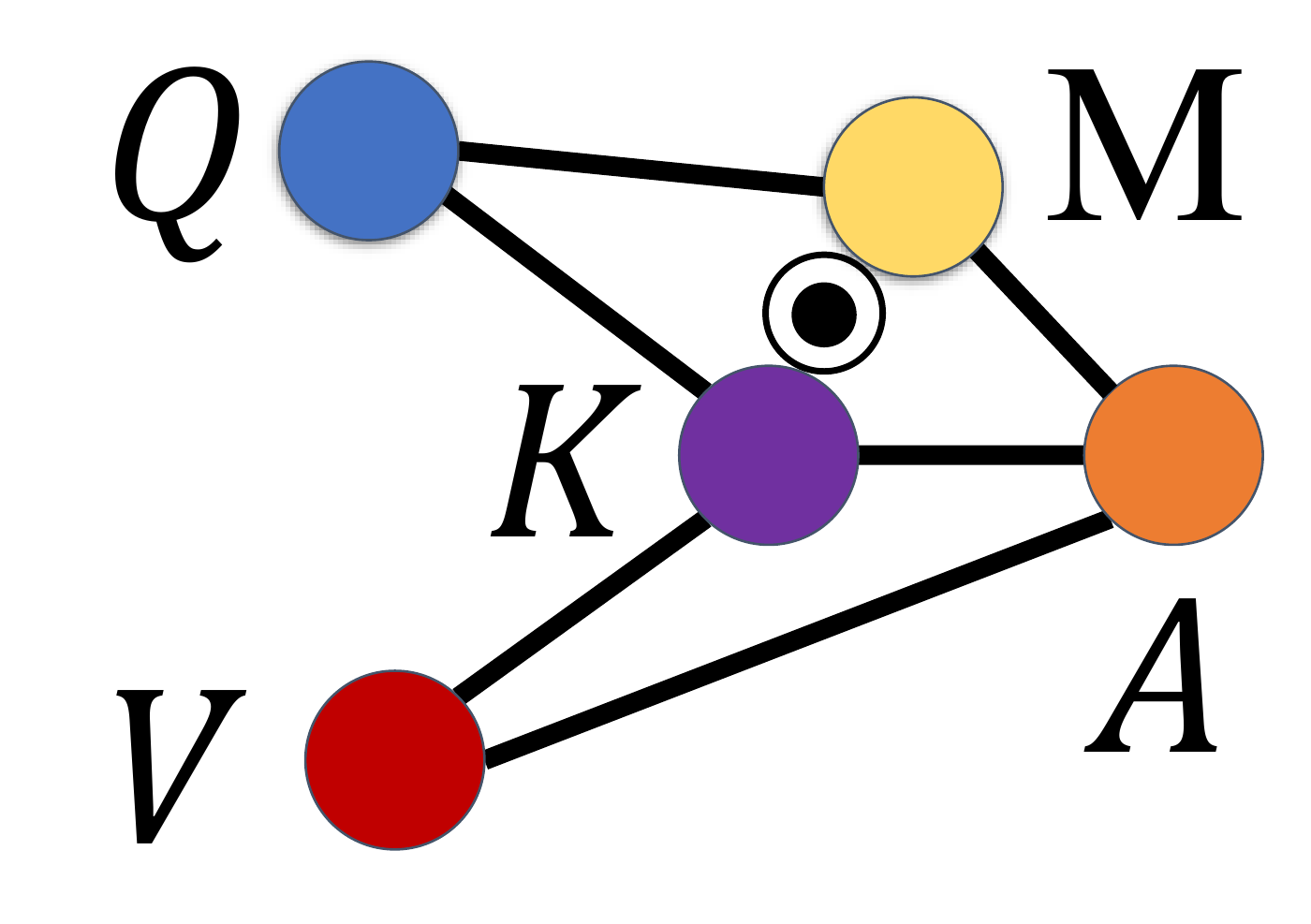}
         \caption{RUBi}
         \label{fig:RUBi_background}
     \end{subfigure}
     \begin{subfigure}[t]{0.22\textwidth}
         \centering
         \includegraphics[width=\textwidth]{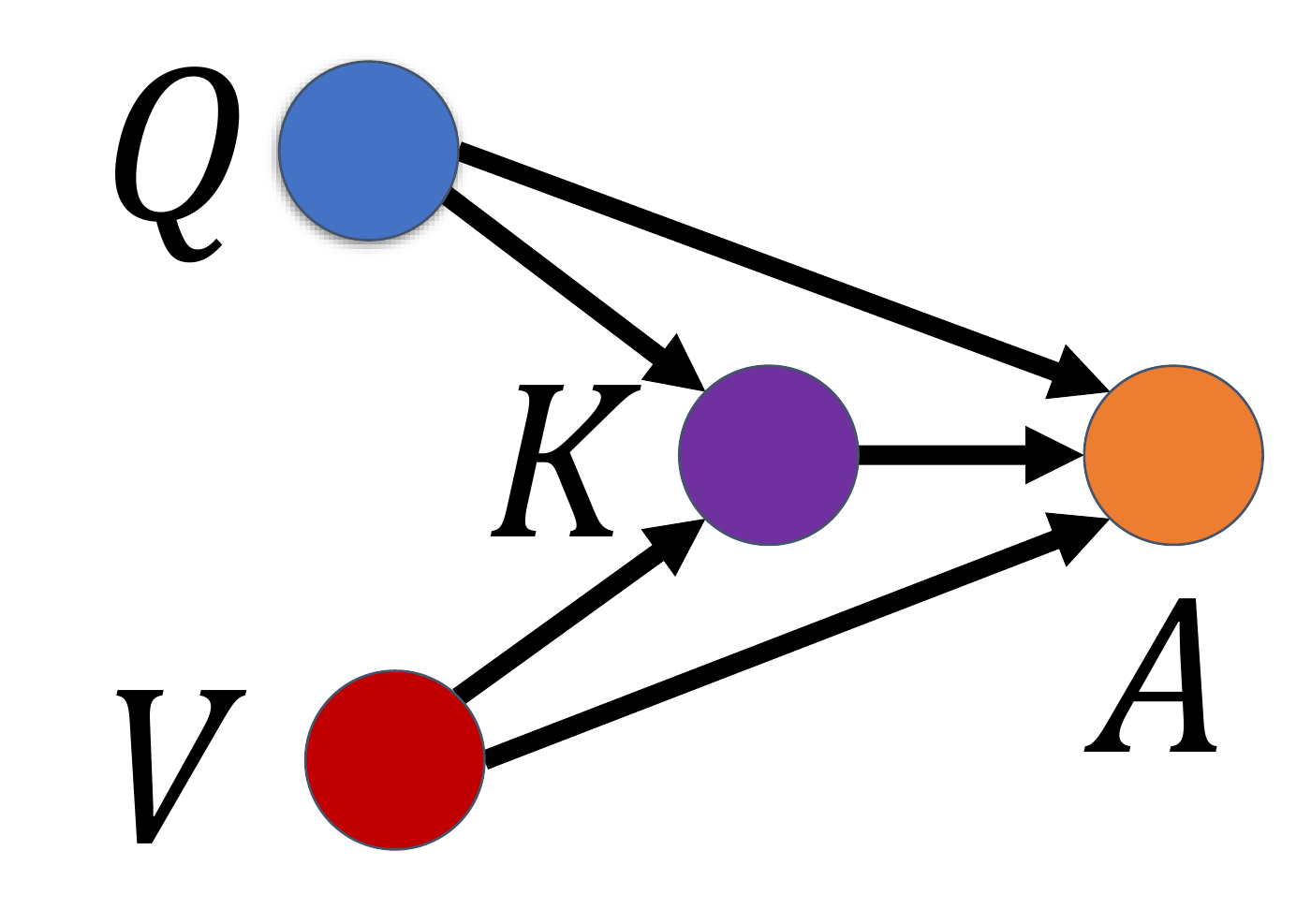}
         
         \caption{CF-VQA graph}    
         \label{fig:CF_VQA_background_a}
     \end{subfigure}
     \begin{subfigure}[t]{0.42\textwidth}
         \centering
        \includegraphics[width=\textwidth]{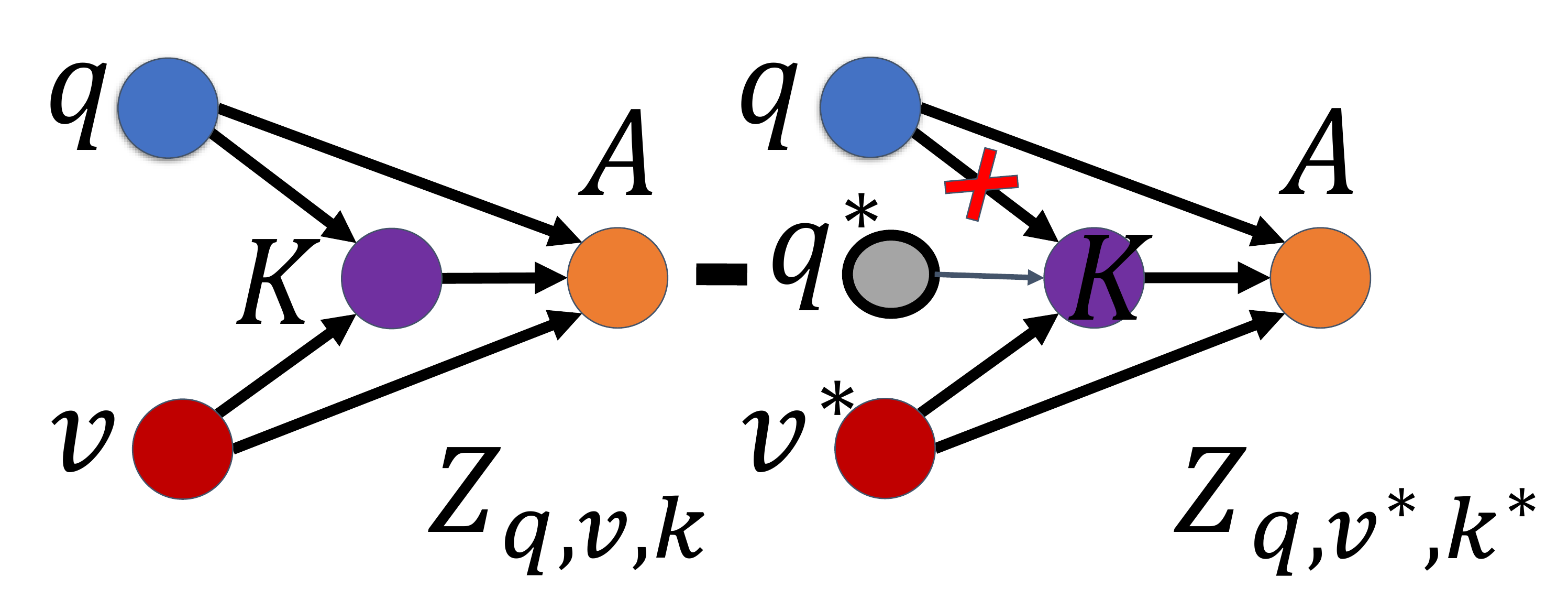}
         
         \caption{Total indirect effect}
         \label{fig:CF_VQA_background_b}
         
     \end{subfigure}
        \caption{VQA graphs related to RUBi and CF-VQA are shown. a) In RUBi, question $Q$ and image $V$ are fused through multimodal knowledge $K$ to obtain an answer $A$, while question-only mask $M$ is applied on $K$; b) causal graph of CF-VQA is shown, where $Q\rightarrow A$ and $V\rightarrow A$ are vision and language shortcuts, all $V$, $Q$, and $K$ are factual; c) output of VQA with counterfactual question $Q=q^*$ and vision $V=v^*$ is subtracted from a regular VQA with factual $V=v$ and $Q=q$.    }
        \label{fig:rubi_cfvqa_background}
\end{figure*}

Our method is motivated by Counterfactual VQA, 
CF-VQA~\cite{niu2021counterfactual}, 
which was motivated by Reducing Unimodal Biases for 
VQA, RUBi~\cite{cadene2019rubi}. 
We review these two methods and their evolution 
in~\ref{subsec:rubi} and~\ref{subsec:cfvqa} and 
then discuss their limitations in~\ref{subsec:limitations}.

\subsection{Reducing Unimodal Biases for VQA}\label{subsec:rubi}

The undirected graph of a RUBi is shown in 
Fig.~\ref{fig:RUBi_background}, with $\{V, Q, K, A, M\}$ 
as set of nodes, $V$: image, $Q$: question, 
$K$: multimodal knowledge, $A$: answer out of 
a set of answers $\mathcal{A}=\{a\}$, $M$: 
question mask. $\mathcal{F}_Q$ is an encoder for questions, 
and $\mathcal{F}_V$ is for images. 
Consequently, a multimodal function $\mathcal{F}_{VQ}$ 
is used to obtain $k=\mathcal{F}_{VQ}(v,q)$. 
An auxiliary neural network $nn_q$ is trained 
to classify answers based on 
only $\{q, a\}$ pairs. 
Then, the classification head is discarded at 
inference to obtain the masks $m=\sigma(nn_q(\mathcal{F}_Q(q)))$, 
where $\sigma$ is the \textit{sigmoid} function.
The masks are then applied to the multimodal classification
$k\odot m$ to reduce the language bias.

\subsection{Counterfactual VQA (CF-VQA)}\label{subsec:cfvqa}

CF-VQA uses counterfactual thinking and causal 
inference to improve RUBi, by only adding one 
learnable parameter. 
The causal graph of CF-VQA is shown in 
Fig.~\ref{fig:CF_VQA_background_a}. 
The graph $\mathcal{G}=\{\mathcal{V},\mathcal{E}\}$ 
is a Directed Acyclic Graph (DAG), 
where $\mathcal{V}=\{V, Q, K, A\}$ 
with a set of causal edges such that 
if $Q \rightarrow K$, then $Q$ is a direct
cause of $K$. 
Moreover, $Q$ is an indirect cause 
of $A$ through the \textit{mediator} $K$, 
as $Q\rightarrow K \rightarrow A$. 
The causal edge assumption states that
every parent is a direct cause of all its
children. 
The answer $a$ can be defined in a multi-class
classifier using logits (score) $Z$.
Therefore, for $h$ as a 
fusion function, for question
$q$, image $v$, and multimodal 
knowledge $k$, these scores for 
question-only, multimodal fused and vision-only 
are:
\begin{equation}\label{eq:archi}
    \begin{gathered}
        Z_q=\mathcal{F}_Q(q),\quad Z_v=\mathcal{F}_V(v),\quad Z_k=\mathcal{F}_{VQ}(v,q),\\
        Z_{q,v,k}=h(Z_q,Z_v,Z_k),
    \end{gathered}
\end{equation}
Denoting answer score $Z_{q,v,k}$ as:
\begin{equation}
    Z_{q,v,k}=Z(Q=q, V=v, K=k),
\end{equation}
the total effect (TE) of $V=v$ and 
$Q=q$ on $A=a$, according to~\cite{niu2021counterfactual}, 
is defined as:
\begin{equation}\label{eq:te}
    TE=Z_{q,v,k}-Z_{q^*,v^*,k^*},
\end{equation}
where $Z_{q^*,v^*,k^*}$ is answer logits $Z$ for 
counterfactual question $q^*$, counterfactual
image $v^*$, and counterfactual multimodal 
knowledge $k^*$. 
The total effect can be decomposed 
into natural direct effect (NDE) and 
total indirect effect (TIE):
\begin{equation}\label{eq:te_decompose}
    TE = TIE + NDE.
\end{equation}
NDE for the question-only branch is
$Q\rightarrow A$ by comparing 
$Z_{q,v^*,k^*}$ and $Z_{q^*,v^*,k^*}$:
\begin{equation}\label{eq:nde}
    NDE = Z_{q,v^*,k^*} - Z_{q^*,v^*,k^*}.
\end{equation}
Finally,
using \eqref{eq:te}, \eqref{eq:te_decompose}, 
and \eqref{eq:nde}, TIE will be:
\begin{equation}
    TIE = Z_{q,v,k} - Z_{q,v^*,k^*},
\end{equation}
as shown in Fig.~\ref{fig:CF_VQA_background_b}.
Consequently, the logits $Z_{q,v,k}$ is parametrized 
as $\mathcal{F}_Q$: $Q\!\rightarrow\!A$, 
and $\mathcal{F}_{VQ}$: $(V,Q)\!\rightarrow\!K\!\rightarrow\!A$. 
The question-only and vision-only 
logits $Z_q$ and $Z_v$ will be as:
\begin{equation}
\begin{split}
    Z_b=
    \begin{cases}
    z_{b}=\mathcal{F}_{B}(b) & \text{ if $B=b$}\\
    z^*_{b}=c & \text{ if $b=\varnothing$}\\
    \end{cases},
\end{split}\label{eq:zq}
\end{equation}
where $B\in \{Q, V\}$, and $c$ as a constant, learnable feature, as described in~\cite{niu2021counterfactual}, and $z^*_{b}$ is a counterfactual realization of $Z_b$. Furthermore, multimodal knowledge's logit $Z_k$ is defined as:
\begin{equation}
\begin{split}
    Z_{k}=
    \begin{cases}
    z_k=\mathcal{F}_{VQ}(v,q) & \text{ if $V=v$ and $Q=q$}\\
    z^*_k=c & \text{ if $V=\varnothing$ or $Q=\varnothing$}\\
    \end{cases}.
\end{split}\label{eq:zk}
\end{equation}

\subsection{Limitations}\label{subsec:limitations}
\noindent\textbf{Visual Bias in VQA:} Visual bias is relatively recent, especially since the language has been known as the primary source of spurious question-answer correlations and has shadowed the research on vision bias~\cite{gat2020removing}. Some recent works have studied the VQA systems' shortcuts directly from the vision's contextual information to the answer~\cite{gupta2022swapmix}. This includes the learning biases of the colors and pixels or the context of the image and a lack of accurate attention to the important parts~\cite{gupta2022swapmix}. We propose a method that mitigates both the language and vision biases in a counterfactual explain-away network that enhances the multimodality of the VQA models.

\noindent\textbf{Memory's Influence:}
Recent studies suggest that memory influences visual perception~\cite{lupyan2020effects}, the problem that we illustrate with a famous example of Rubin's face~\cite{rubin1915synsoplevede}, as shown in Fig.~\ref{fig:teaser_pwvqa}, where the same image can be perceived differently. Rubin discussed this as memory bias~\cite{rubin1915synsoplevede, rubin1921visuell}, which accumulates based on the experiences of individuals. The importance of visual perception may depend on the language~\cite{lupyan2020effects}, location~\cite{zhang2021situatedqa}, time~\cite{zhang2021situatedqa}, and experiences of people~\cite{liu2021visually}, which may lead to interpret an image in different ways. Although prior works on other datasets have tried to add concepts or new languages~\cite{liu2021visually}, our work proposes a method to address the experience of the annotator as an unobserved confounder to reduce experience bias.

\section{Possible Worlds VQA (PW-VQA)}\label{sec:pwvqa}
In this section, we explain the
proposed method in four subsections. 
First, we simultaneously model the
language and vision bias using a causal view.
Then we model experience bias as unobserved 
confounders of the VQA systems. 
Third, a counterfactual method is proposed 
in the subsequent subsection to solve these problems. 
Finally, we propose a novel strategy to fuse 
multimodal vision and language information in VQA systems.

\begin{figure}[!h]
    \centering
    \includegraphics[width=0.7\linewidth]{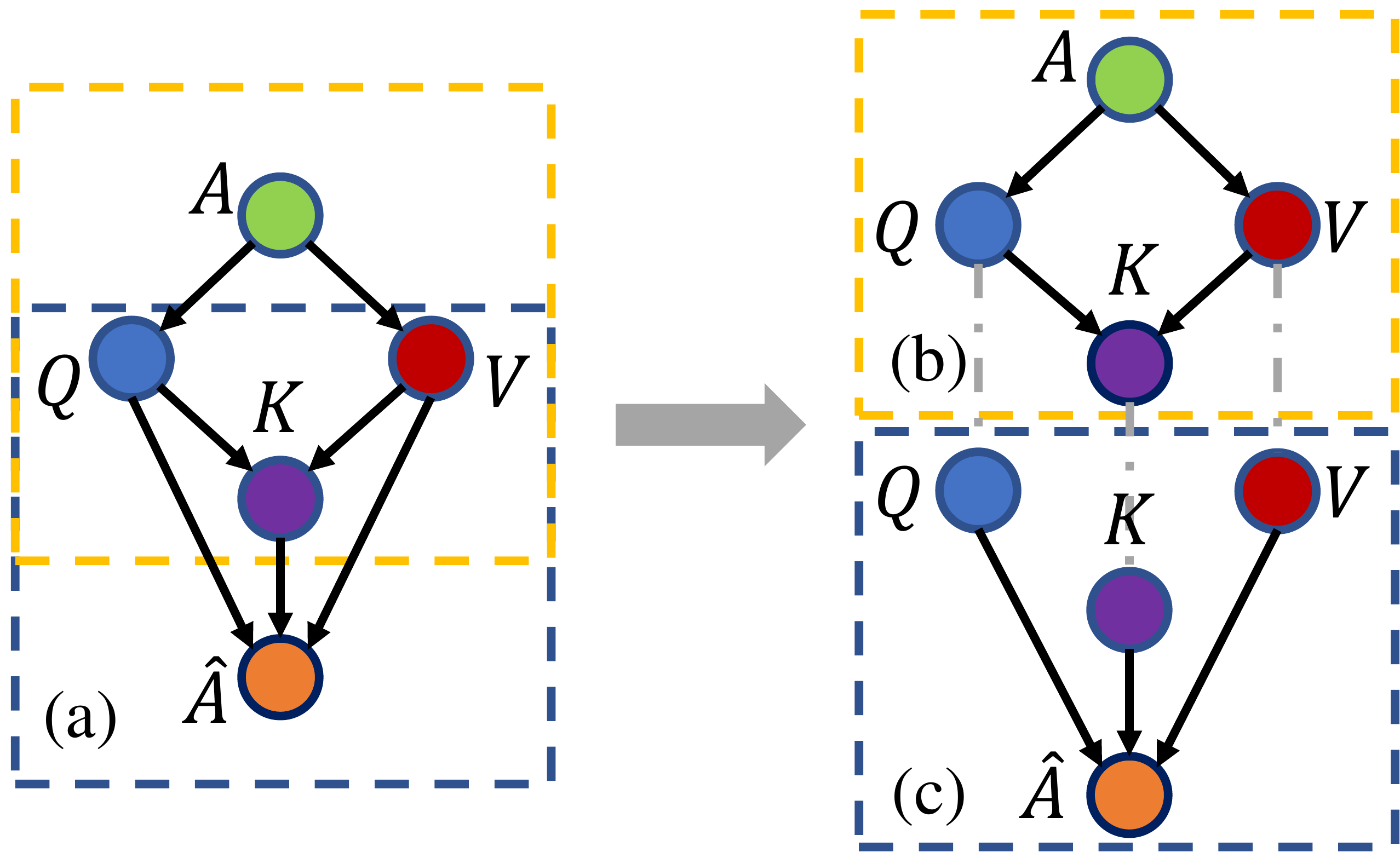}
    \caption{The proposed causal graph reformulates the VQA problem by stating that a) the answer $A$ is a cause of the question $Q$, and vision $V$, and the final estimated answer $\hat{A}$ is achieved by fusing $V$ and $Q$ information. b) The anticausal subgraph consists of the ground-truth answer $A$ that is a cause of the $V$ and $Q$, which leads to multimodal knowledge $K$. c) The collider $Q\rightarrow K \leftarrow V$ is an explain-away network that models the language-vision bias.}
    \label{fig:explain-away}
\end{figure}

Assume that a multimodal knowledge $K$ contains
fused information of question $Q$ and vision $V$ used in a VQA system. 
We propose the causal graph
$\mathcal{G}= \{\mathcal{V}, \mathcal{E}\}$ 
with the set of nodes $\mathcal{V} = \{Q, V, K, A, \hat{A}\}$, 
which is shown in Fig.~\ref{fig:explain-away}a to model VQA systems. 

\begin{figure}
    \centering
    \includegraphics[width=0.4\linewidth]{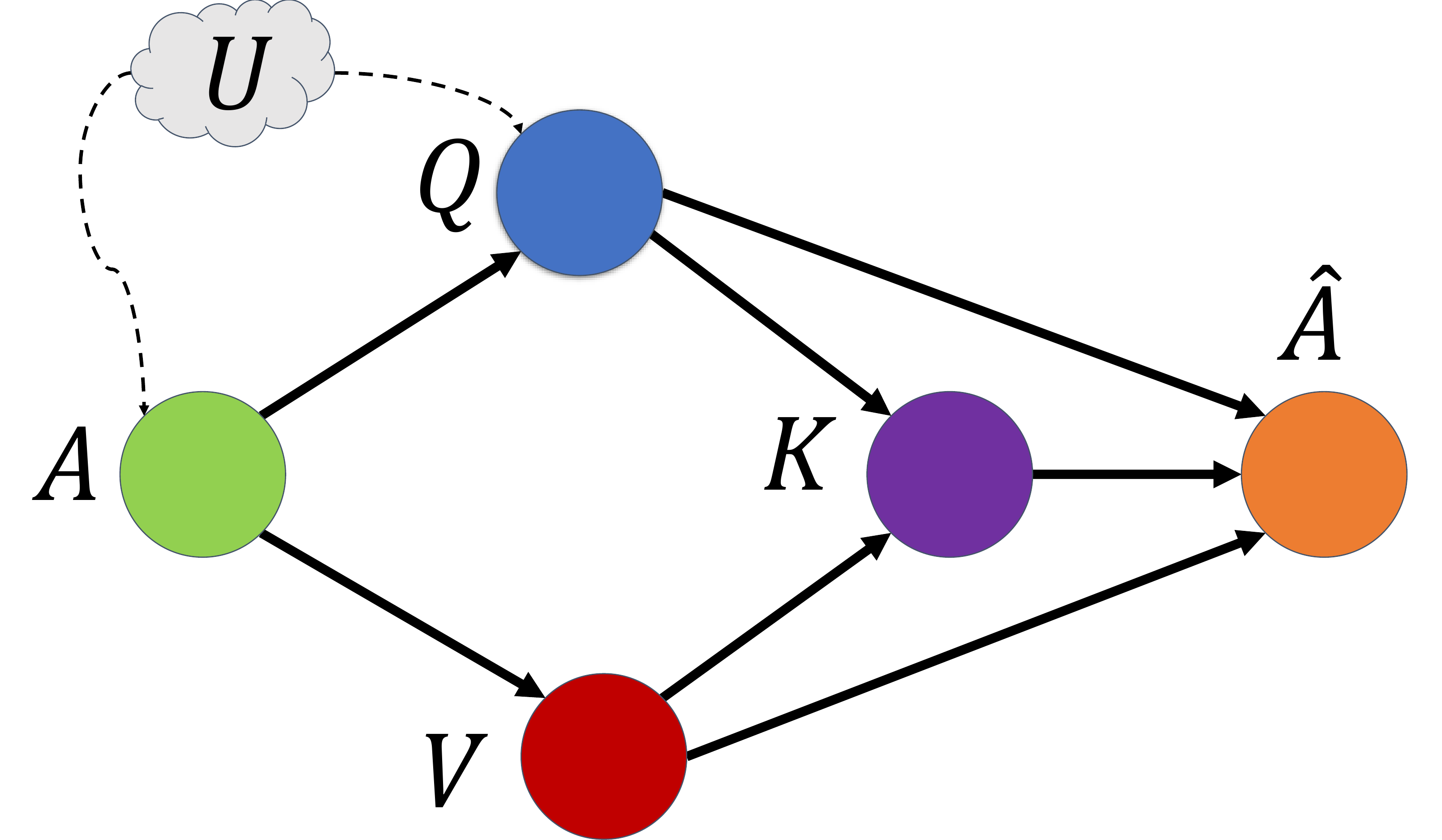}
    \caption{The causal graph of the VQA where the question $Q$ and the answer $A$ are influenced by unobserved confounder $U$.   }
    \vspace{-.2in}
    \label{fig:full_graph_proposed}
\end{figure}

Inspired by the anticausal 
learning~\cite{janzing2012causal, arjovsky2019invariant}, 
we model the answer $A$ as a cause
of both the images $V$ and question $Q$. 
Unlike previous works 
\cite{niu2021counterfactual, cadene2019rubi, niu2021introspective}, 
we distinguish between the ground-truth 
answer $A$ for the training of the VQA model 
and the estimated answer $\hat{A}$ when the
model is used in practice (test). 
Therefore, as shown in Fig.~\ref{fig:explain-away}b, $Q$ 
and $V$ have a causal effect on $K$ and are 
also a child of the answer $ A $. \newline
\textbf{Collider Confounder in Vision and Language:} 
The relationship $ Q \rightarrow A $ creates 
a spurious correlation between the question
$Q$ directly to the answer $A$. 
Therefore, the $ V \rightarrow K \rightarrow A $ 
information is ignored. Contrarily the VQA
models may shortcut visual information to
answer $ V \rightarrow K \rightarrow A $ rather 
than multimodal knowledge~\cite{gupta2022swapmix}. 
By looking at the subgraph shown in 
Fig.~\ref{fig:explain-away}c, the explain-away network, 
or collider bias network simultaneously can
model vision and language bias. 
The relationship $Q \rightarrow \hat{A} \leftarrow V$ is 
a collider, a primitive graph structure, \textit{aka} 
explain-away network. 
Consequently, having a strong connection 
$Q \rightarrow \hat{A}$ removes the 
dependency of the $\hat{A}$ on $V$. 
Noteworthy that there are useful information 
and harmful biases in both vision and language. 
Our explain-away method aims to remove 
biases but keep good information. 
\begin{figure}
    \centering
    \includegraphics[width=0.7\linewidth]{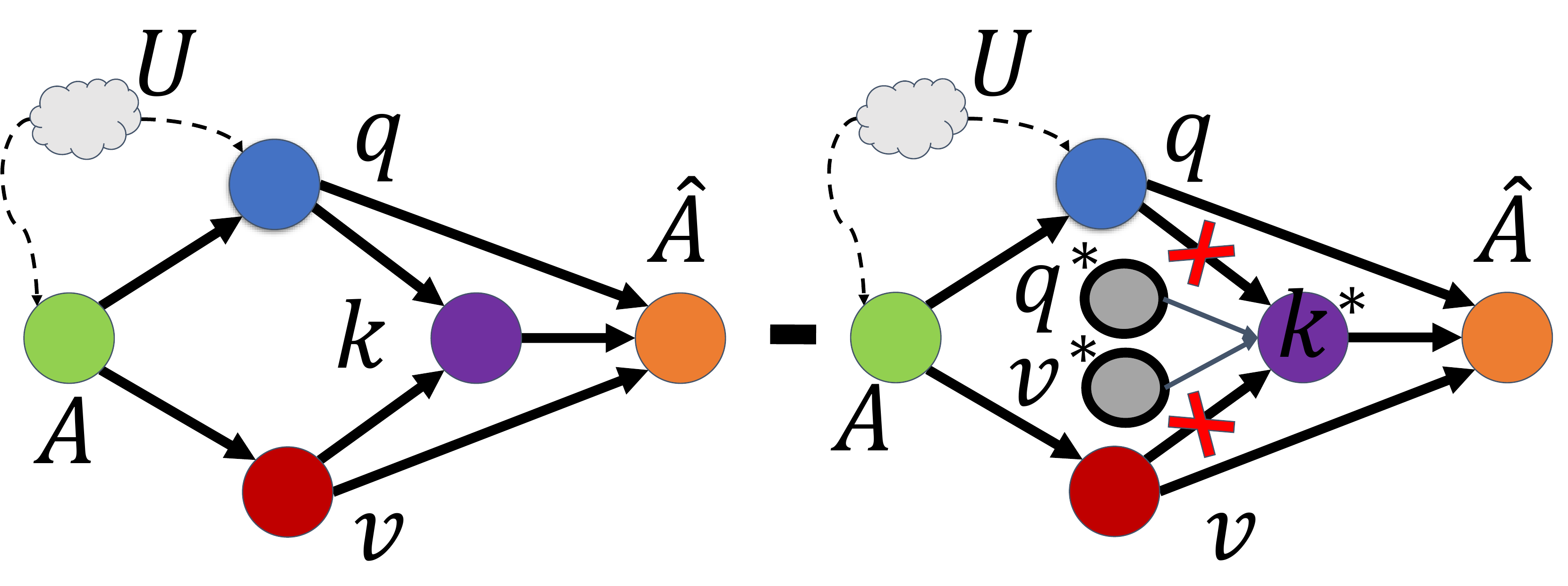}
    \caption{The multimodal knowledge $K=k^*$ is counterfactual, while Q and V are facts ($Q=q, V=v, K=k^*$), then, the natural indirect effect (NDE) is subtracted from the total effect (TE) to obtain total indirect effect (TIE). The values $V\!=\!v$ and $Q\!=\!q$ are fact, and  $V\!=\!v^*$ and $Q\!=\!q^*$, which leads to $K\!=\!k^*$ are counterfactuals. }
    \label{fig:proposed_technique_graphs}
\end{figure}
Therefore we introduce the collider 
of $Q \rightarrow \hat{A} \leftarrow V$  as 
a source of vision-language bias in VQA models. \newline
\textbf{Experience as an Unobserved Confounder:} 
Based on the proposed causal graph $\mathcal{G}$, 
a novel source of confounding is introduced related 
to the experience of the annotator that happens
during the preparation of the datasets. 
As an example of experience bias, we have seen the visual
illusion problem in Fig.~\ref{fig:teaser_pwvqa}. 
To be specific, selecting questions $Q$ and 
answering $A$ to the question based on an 
image $V$ relies on the personal preferences of the annotator. 
Therefore, unobserved bias $U$ depends 
on the personal preferences of the annotator.
The proposed causal graph for the VQA models 
with unobserved confounder is shown 
in Fig.~\ref{fig:full_graph_proposed}. 
Consequently, by looking into different 
paths that are parents or ancestors 
of $\hat{A}$, they can be listed as 
$U \rightarrow A \rightarrow Q \rightarrow K \rightarrow \hat{A}$,
$U \rightarrow A \rightarrow Q \rightarrow K \rightarrow \hat{A}$,
$U \rightarrow  A \rightarrow V \rightarrow \hat{A}$,
and  $U \rightarrow A \rightarrow V \rightarrow K \rightarrow \hat{A}$.
The same can be listed for $U \rightarrow Q$ 
paths; however, only $K \rightarrow \hat{A}$ 
is of interest for the VQA models.\newline
\textbf{Explain-Away Fusion Strategy (EA):}
We propose the following Explain-Away (EA) fusion function as follows.
For parametrization, we use similar notations as~\cite{niu2021counterfactual}. Therefore, the score $Z_{q,v,k}$ which is the feature space of the fusion $K$, is {parametrized} as $\mathcal{F}_Q$: $Q\!\rightarrow\!\hat{A}$, and $\mathcal{F}_{VQ}$: $(V,Q)\!\rightarrow\!K\!\rightarrow\!\hat{A}$. 

Based on $Z_q,Z_v$, and $Z_k$, we define the fusion function as follows:
\begin{flalign}\label{eq:fusion-proposed}
\text{(EA)}&&h(Z_q,Z_v,Z_k)=\frac{1}{\alpha + 1}\log (Z_{\text{EA}}),&&
\end{flalign}
where $Z_{\text{EA}}$ is defined as:
\begin{equation}
\begin{split}
  Z_{\text{EA}}=
  & \sigma(Z_q)^{\alpha}\sigma(Z_v)^{\alpha+1}\sigma(Z_k)^{\alpha+1} \\ &+\sigma(Z_q)^{\alpha+1}\sigma(Z_v)^{\alpha}\sigma(Z_k)^{\alpha+1}\\ &+\sigma(Z_q)^{\alpha+1}\sigma(Z_v)^{\alpha+1}\sigma(Z_k)^{\alpha},
\end{split}
\end{equation}
and $\alpha \ge 0$ is a free parameter that can be defined based on empirical analysis.
\newline
\textbf{Unobserved Confounding Bias Reduction:} Since the model relies on the fused information $K$ of $V$ and $Q$, and as shown in Fig.~\ref{fig:proposed_technique_graphs}, the confounding bias of vision-language can be removed by maximizing the total indirect effect (TIE) by subtracting natural direct effect (NDE) of this confounding influence from its total effect (TE)~\cite{Pearl2001DirectAI}:
\begin{equation}\label{eq:TIE_TE}
\begin{split}
\TIE&=\TE-\NDE\\
&= h(Z_q,Z_v,Z_k)-h(Z_q,Z_v,Z_{k^*}),
\end{split}
\end{equation}
where $K^*$ is a counterfactual of $K$, as described in~\eqref{eq:zk}. 
As the influence of the unobserved confounding bias is subtracted in~\eqref{eq:TIE_TE}, it will block the influence of the explain-way of vision-language and experience biases altogether. 
By blocking the two paths $V\rightarrow K$ and $Q\rightarrow K$, all influences from unobserved confounding bias are blocked. 

\noindent \textbf{Training:} For the 
training of the network, we use vision-only 
branch $\mathcal{L}_{V\!A}(v,a)$, 
question-only branch $\mathcal{L}_{Q\!A}(q,a)$, 
and multimodal fusion branch $\mathcal{L}_{V\!Q\!A}(v,q,a)$. As illustrated in Fig.~\ref{fig:full_graph_proposed}, given a triplet $(v,q,a)$ where $a$ is the ground-truth answer of image-question pair $(v,q)$, the branches are optimized by minimizing the cross-entropy losses over the scores $Z_{q,v,k}$, $Z_q$ and $Z_v$:~\cite{niu2021counterfactual}:
\begin{equation}\label{eq:loss}
    \mathcal{L}_{cls}=\mathcal{L}_{V\!Q\!A}(v,q,a)+\mathcal{L}_{Q\!A}(q,a)+\mathcal{L}_{V\!A}(v,a),
\end{equation}
where $\mathcal{L}_{V\!Q\!A}$, $\mathcal{L}_{Q\!A}$ and $\mathcal{L}_{V\!A}$ are over $Z_{q,v,k}$, $Z_{q}$ and $Z_{v}$. A learnable parameter $c$ in Eq.~\eqref{eq:zq}-\eqref{eq:zk}, which controls the sharpness of the distribution of $Z_{q,v^*,k^*}$ is also included, as the sharpness of NDE should be similar to that of TE~\cite{hinton2015distilling, niu2021counterfactual}. An improper $c$ would lead to the domination of TIE in Eq.~\eqref{eq:TIE_TE} by either TE or NDE. Thus, we use KL-divergence to estimate $c$:
\begin{equation}
    \mathcal{L}_{kl}=\frac{1}{|\mathcal{A}|}\sum_{a\in\mathcal{A}}-p(a|q,v,k)\log p(a|q,v^*,k^*),
\end{equation}
where $p(a|q,v,k)\!=\!\softmax(Z_{q,v,k})$ and $p(a|q,v^*,k^*)\!=\!\softmax(Z_{q,v^*,k^*})$. Only $c$ is updated when minimizing $\mathcal{L}_{kl}$. 
The final loss is
the combination of $\mathcal{L}_{cls}$ and $\mathcal{L}_{kl}$:
\begin{equation}
    \mathcal{L}_{final}=\sum_{(v,q,a)\in\mathcal{D}}\mathcal{L}_{cls}+\mathcal{L}_{kl}
\end{equation}

\noindent \textbf{Inference}. For the inference, we use the debiased causal effect for inference, which is implemented as:
\begin{equation}\label{eq:inf}
\begin{split}
\TIE=\TE-\NDE&=Z_{q,v,k}-Z_{q,v^*,k^*}\\
&=h(z_q,z_v,z_k)-h(z_q,z^*_v,z^*_k).
\end{split}
\end{equation}

\begin{table*}[!t]
\centering
\caption{The table lists the accuracy values for the most recent studies, especially on both VQA-CP v2 and VQA v2 datasets. We show the best-performing method with bold and the second-best-performing method with an underline. We use a dash for the papers that miss reporting performance values on datasets.}
\label{tab:all}
\vspace{-2mm}
\scalebox{0.85}{
\begin{tabular}{|llrrrrrrrr|}
\hline
Test set                                                         &       & \multicolumn{4}{|l|}{VQA-CP v2 test}                                                                       & \multicolumn{4}{|l|}{VQA v2 test}                                                                         \\ \cline{3-10} 
Methods                                                          & Base  & \multicolumn{1}{|l|}{All} & \multicolumn{1}{|l|}{Y/N} & \multicolumn{1}{|l|}{Num.} & \multicolumn{1}{|l|}{Other} & \multicolumn{1}{|l|}{All} & \multicolumn{1}{|l|}{Y/N} & \multicolumn{1}{|l|}{Num.} & \multicolumn{1}{|l|}{Other} \\ \hline
GVQA~\citep{agrawal2018don}                & -     & 31.30                    & 57.99                   & 13.68                    & 22.14                     & 42.24                   & 72.03                   & 31.17                    & 34.65                     \\
SAN~\citep{yang2016stacked}                & -     & 24.96                   & 38.35                   & 11.14                    & 21.74                     & 52.41                   & 70.06                   & 39.28                    & 47.84                     \\
UpDn~\citep{anderson2018bottom}            & -     & 39.74                   & 42.27                   & 11.93                    & 46.05                     & {63.48}                   & 81.18                   & 42.14                    & \textbf{55.66}                     \\
S-MRL~\citep{cadene2019rubi}               & -     & 38.46                   & 42.85                   & 12.81                    & 43.20                      & 63.10                    & \multicolumn{1}{l}{-}   & \multicolumn{1}{l}{-}    & \multicolumn{1}{l|}{-}     \\ \hline
\multicolumn{10}{|l|}{{\color[HTML]{EA4335} \textit{Methods based on modifying language module or using language prior:}}}                                                                                                                                                                       \\ \hline
DLR~\citep{jing2020overcoming}             & UpDn  & 48.87                   & 70.99                   & 18.72                    & 45.57                     & 57.96                   & 76.82                   & 39.33                    & 48.54                     \\
VGQE~\citep{gouthaman2020reducing}         & UpDn  & 48.75                   & \multicolumn{1}{l}{-}   & \multicolumn{1}{l}{-}    & \multicolumn{1}{l}{-}     & \textbf{64.04}                   & \multicolumn{1}{l}{-}   & \multicolumn{1}{l}{-}    & \multicolumn{1}{l|}{-}     \\
VGQE~\citep{gouthaman2020reducing}         & S-MRL & 50.11                   & 66.35                   & 27.08                    & 46.77                     & 63.18                   & \multicolumn{1}{l}{-}   & \multicolumn{1}{l}{-}    & \multicolumn{1}{l|}{-}     \\
AdvReg.    ~\citep{ramakrishnan2018overcoming}   & UpDn  & 41.17                   & 65.49                   & 15.48                    & 35.48                     & 62.75                   & 79.84                   & 42.35                    & 55.16                     \\
RUBi~\citep{cadene2019rubi}                & UpDn  & 44.23                   & 67.05                   & 17.48                    & 39.61                     & \multicolumn{1}{l}{-}   & \multicolumn{1}{l}{-}   & \multicolumn{1}{l}{-}    & \multicolumn{1}{l|}{-}     \\
RUBi~\citep{cadene2019rubi}                & S-MRL & 47.11                   & 68.65                   & 20.28                    & 43.18                     & 61.16                   & \multicolumn{1}{l}{-}   & \multicolumn{1}{l}{-}    & \multicolumn{1}{l|}{-}     \\
LM~\citep{clark2019don}                    & UpDn  & 48.78                   & 72.78                   & 14.61                    & 45.58                     & 63.26                   & 81.16                   & 42.22                    & 55.22                     \\
LM+H~\citep{clark2019don}                  & UpDn  & 52.01                   & 72.58                   & 31.12                    & 46.97                     & 56.35                   & 65.06                   & 37.63                    & 54.69                     \\
CF-VQA (SUM)~\citep{niu2021counterfactual} & UpDn  & 53.55                   & \textbf{91.15}                   & 13.03                    & 44.97                     & \underline{63.54}                   & \textbf{82.51}                   & \textbf{43.96}                    & 54.30                      \\
CF-VQA (SUM)~\citep{niu2021counterfactual} & S-MRL & 55.05                   & \underline{90.61}                   & 21.50                     & 45.61                     & 60.94                   & 81.13                   & 43.86                    & 50.11                     \\
GGE-DQ-tog~\citep{han2021greedy}           & UpDn  & 57.32                   & 87.04                   & 27.75                    & \textbf{49.59}                     & 59.11                   & 73.27                   & 39.99                    & 54.39                     \\ \hline
\multicolumn{10}{|l|}{{\color[HTML]{EA4335} Methods based on reducing visual bias or enhancing visual attention/grounding:}}                                                                                                                                                                     \\ \hline
AttAlign ~\citep{selvaraju2019taking}            & UpDn  & 39.37                   & 43.02                   & 11.89                    & 45.00                        & 63.24                   & 80.99                   & 42.55                    & 55.22                     \\
HINT   ~\citep{selvaraju2019taking}              & UpDn  & 46.73                   & 67.27                   & 10.61                    & 45.88                     & 63.38                   & 81.18                   & 42.99                    & \underline{55.56}                     \\
SCR~\citep{wu2019self}                     & UpDn  & 49.45                   & 72.36                   & 10.93                    & \underline{48.02}                     & 62.20                    & 78.80                    & 41.60                     & 54.50                      \\ \hline
\multicolumn{10}{|l|}{{\color[HTML]{FF0000} \textit{Methods mitigating both language and vision:}}}                                                                                                                                                                                               \\ \hline
LMH+Fisher~\citep{gat2020removing}         & UpDn  & 54.55                   & 74.03                   & 49.16                    & 45.82                     & \multicolumn{1}{l}{-}   & \multicolumn{1}{l}{-}   & \multicolumn{1}{l}{-}    & \multicolumn{1}{l|}{-}     \\
PW-VQA (ours)                                                     & UpDn  & \underline{59.06}                   & 88.26                   & \underline{52.89}                    & 45.45                     & 62.63                   & \underline{81.80}                   & \underline{43.90}                    & 53.01                     \\
PW-VQA (ours)                                                     & S-MRL & \textbf{60.26}                   & 88.09                   & \textbf{59.13}                    & 45.99                     & 61.25                   & 80.32                   & 43.17                    & 51.53                     \\ \hline
\multicolumn{10}{|l|}{{\color[HTML]{EA4335} \textit{Methods that synthesize data to augment and balance training splits:}}}                                                                                                                                                                      \\ \hline
CVL~\citep{abbasnejad2020counterfactual}    & UpDn  & 42.12                   & 45.72                   & 12.45                    & 48.34                     & \multicolumn{1}{l}{-}   & \multicolumn{1}{l}{-}   & \multicolumn{1}{l}{-}    & \multicolumn{1}{l|}{-}     \\
Unshuffling~\citep{teney2020unshuffling}    & UpDn  & 42.39                   & 47.72                   & 14.43                    & 47.24                     & 68.08                   & 78.32                   & 42.16                    & 52.81                     \\
RandImg~\citep{teney2020value}              & UpDn  & 55.37                   & 83.89                   & 41.60                     & 44.20                      & 57.24                   & 76.53                   & 33.87                    & 48.57                     \\
SSL~\citep{zhu2020overcoming}               & UpDn  & 57.59                   & 86.53                   & 29.87                    & 50.03                     & 63.73                   & \multicolumn{1}{l}{-}   & \multicolumn{1}{l}{-}    & \multicolumn{1}{l|}{-}     \\
CSS~~\citep{chen2020counterfactual}          & UpDn  & 58.95                   & 84.37                   & 49.42                    & 48.21                     & 59.91                   & 73.25                   & 39.77                    & 55.11                     \\
CSS+CL~\citep{liang2020learning}            & UpDn  & 59.18                   & 86.99                   & 49.89                    & 47.16                     & 57.29                   & 67.27                   & 38.40                     & 54.71                     \\
Mutant~~\citep{gokhale2020mutant}            & UpDn  & 61.72                   & 88.90                    & 49.68                    & 50.78                     & 62.56                   & 82.07                   & 42.52                    & 53.28                     \\
LMH+ECD~\citep{kolling2022efficient}       & UpDn  & 59.92                   & 83.23                   & 52.59                    & 49.71                     & 57.38                   & 69.06                   & 35.74                    & 54.25                     \\ \hline
\end{tabular}
}
\end{table*}

\begin{figure*}[!h]
     \centering
     \begin{subfigure}[b]{0.31\textwidth}
         \centering
         \includegraphics[width=\textwidth]{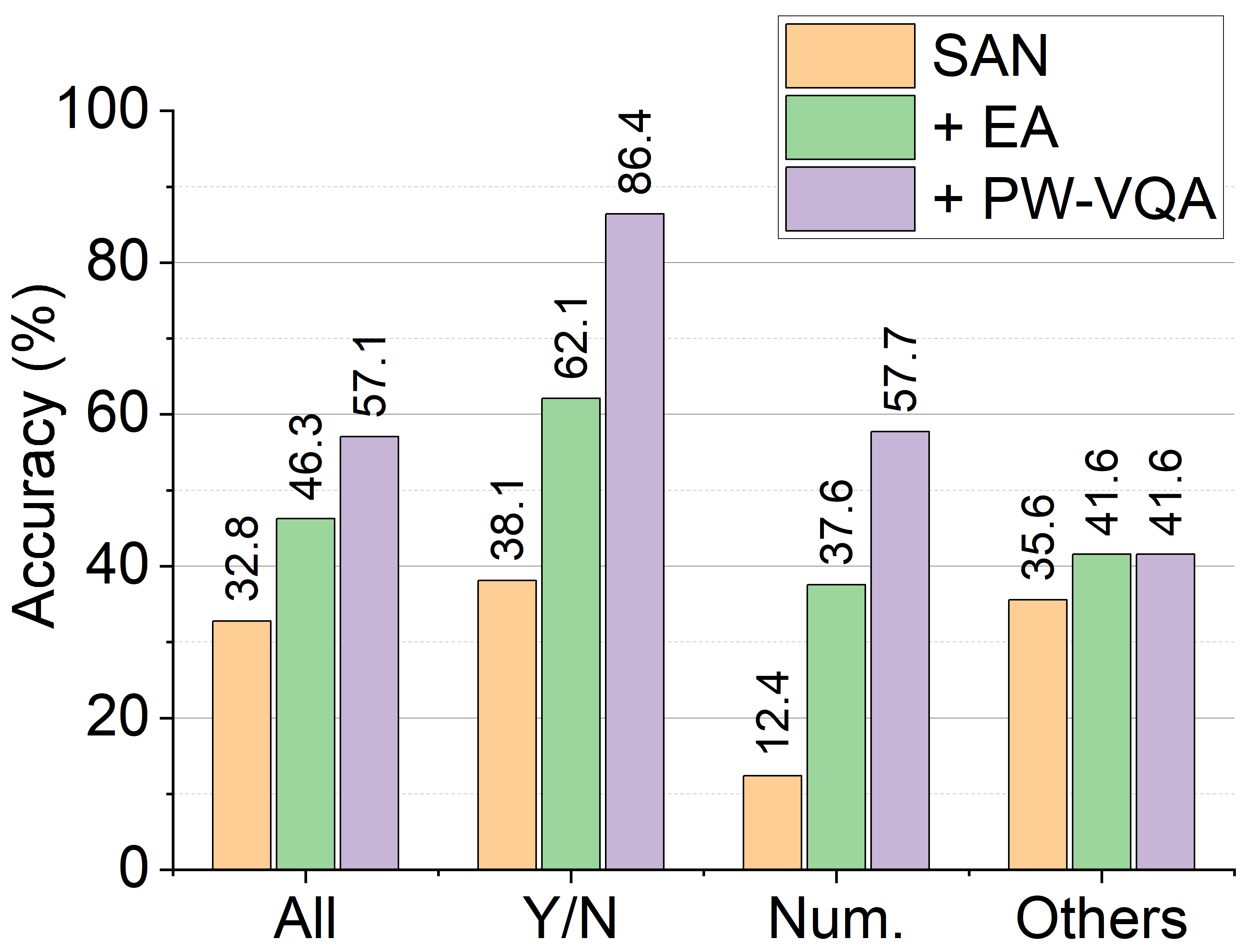}
         \label{fig:SAN_DEP_PWVQA}
     \end{subfigure}
     \hfill
     \begin{subfigure}[b]{0.31\textwidth}
         \centering
         \includegraphics[width=\textwidth]{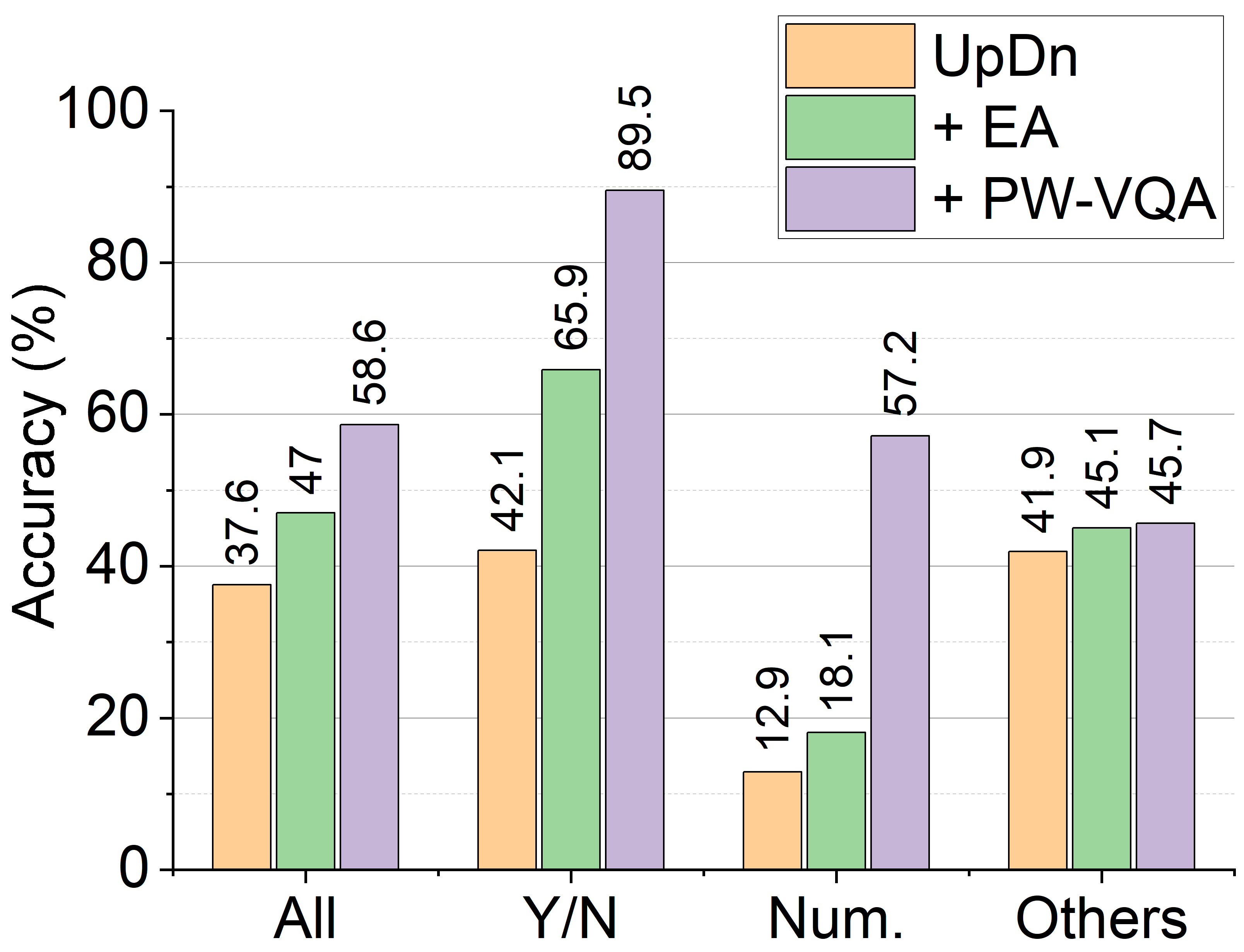}
         \label{fig:UPDN_DEP_PWVQA}
     \end{subfigure}
     \hfill
     \begin{subfigure}[b]{0.31\textwidth}
         \centering
         \includegraphics[width=\textwidth]{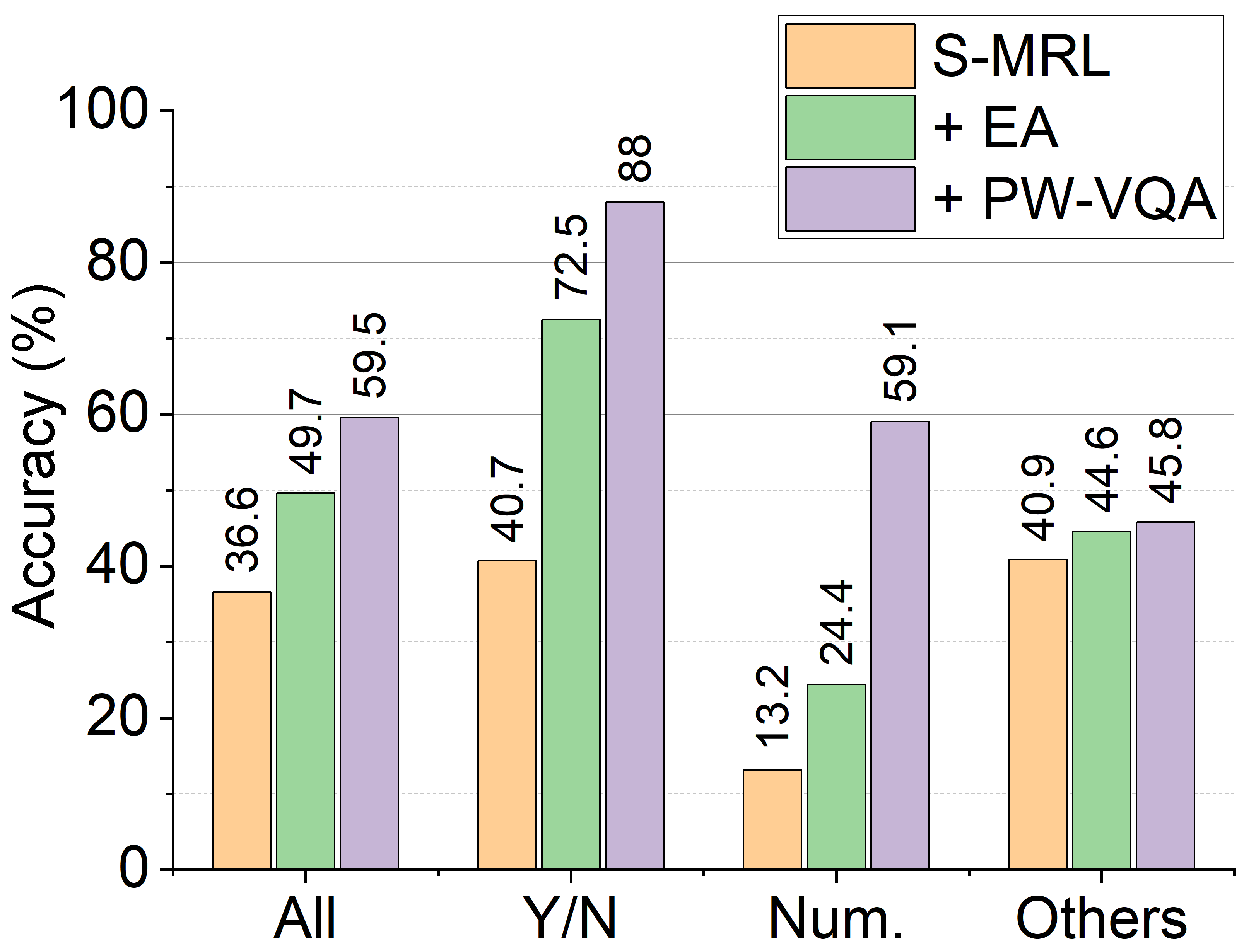}
         \label{fig:SMRL_DEP_PWVQA}
     \end{subfigure}
     \vspace{-.2in}
        \caption{The plots show the backbones using our proposed causal framework (PW-VQA) and fusion strategy (EA). The results are consistently improving for all three different backbones, namely, SAN~\cite{yang2016stacked}, S-MRL~\cite{cadene2019rubi}, and UpDn~\cite{anderson2018bottom}.  }
        \vspace{-.2in}
        \label{fig:backbone_ablaton}
\end{figure*}
\section{Experiments}\label{sec:experiments}

The model can be trained on a computer with a single GeForce GTX 1080 GPU. We used GTX 1080 Ti and RTX A6000 GPUs in our simulations. We used the VQA-CP v2 dataset, which has about 438K questions on the train set and 220K questions on the test set, with corresponding question-answer pairs in the English language~\cite{agrawal2018don}. We applied our VQA model on three backbones, namely Stacked Attention Network ({SAN})~\cite{yang2016stacked}, Bottom-up and Top-down Attention ({UpDn})~\cite{anderson2018bottom}, and a simplified MUREL~\cite{cadene2019murel} ({S-MRL})~\cite{cadene2019rubi}.  Training of our model on VQA-CP v2 dataset~\cite{agrawal2018don} with SAN~\cite{yang2016stacked} takes about 8 hours, and with S-MRL~\cite{cadene2019rubi} and UpDn~\cite{anderson2018bottom} on average takes about 3 hours. The validation on the test split of the VQA-CP v2 dataset takes about 10 minutes. We used accuracy as the evaluation metric. We manually searched the hyperparameter, and we reported those which have the best results in the ablation study. We used a batch size of 256 and 22 epochs for all runs. Increasing the number of epochs does not improve the results since the model converges to a stable result within 22 epochs. We observed that the model does not converge with $\alpha < 1$ values, and therefore we bound $\alpha \ge 1$. We tried $\alpha$ values between 1 to 2 for 11 times and based on empirical study, $\alpha = 1.5$ achieves the best-performing result.\newline
\textbf{Quantitative results:} To compare our method with the available literature on the benchmark datasets, we list the performance values in Table~\ref{tab:all}. Then, to compare reasonably with the existing methods, we divide them into four categories:
1) Methods like DLR~\cite{jing2020overcoming}, VGQE~\cite{gouthaman2020reducing}), AdvReg~\cite{ramakrishnan2018overcoming}, RUBi~\cite{cadene2019rubi}, LM~\cite{clark2019don}, LM+H~\cite{clark2019don}, CF-VQA\cite{niu2021counterfactual}, GGE-DQ-tog~\cite{han2021greedy} modify language modules or use language before suppress, control, or mask language shortcuts. However, these methods only consider spurious language correlations and neglect vision in their schema.
2) Some approaches, such as AttAlign~\cite{selvaraju2019taking}, {HINT}~\cite{selvaraju2019taking}, {SCR}~\cite{wu2019self}  mitigate visual biases by loosening contextual ties to the answer or improving visual grounding and attention via human feedback, de-coupling shortcuts that couple vision to answer. 
3) Other approaches like LMH+Fisher~\cite{gat2020removing} mitigate both language and vision bias together, attempting to balance two modalities of vision and language for robust multimodal inference.  Our proposed method here is in this class.
4) Methods such as {CVL}~\cite{abbasnejad2020counterfactual}, {Unshuffling}~\cite{teney2020unshuffling},  {RandImg}~\cite{teney2020value}, {SSL}~\cite{zhu2020overcoming}, {Mutant}~\cite{gokhale2020mutant}, {CSS}~\cite{chen2020counterfactual}, {CSS+CL}~\cite{liang2020learning},  LMH+ECD~\cite{kolling2022efficient} synthesize samples and augment data to balance training and test sets; however, it violates the main idea of the VQA-CP v2 dataset. Therefore, it is not fair to compare them with our method; however, we include them in our results for inclusiveness. 

\begin{figure*}[t]
     \centering
     \begin{subfigure}[b]{0.49\textwidth}
         \centering
         \includegraphics[width=\textwidth]{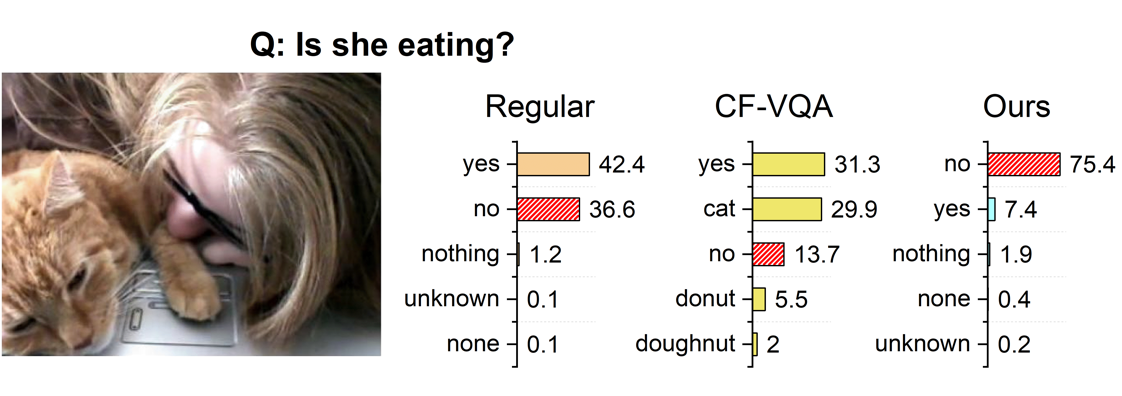}
     \end{subfigure}
     \hfill
     \begin{subfigure}[b]{0.49\textwidth}
         \centering
         \includegraphics[width=\textwidth]{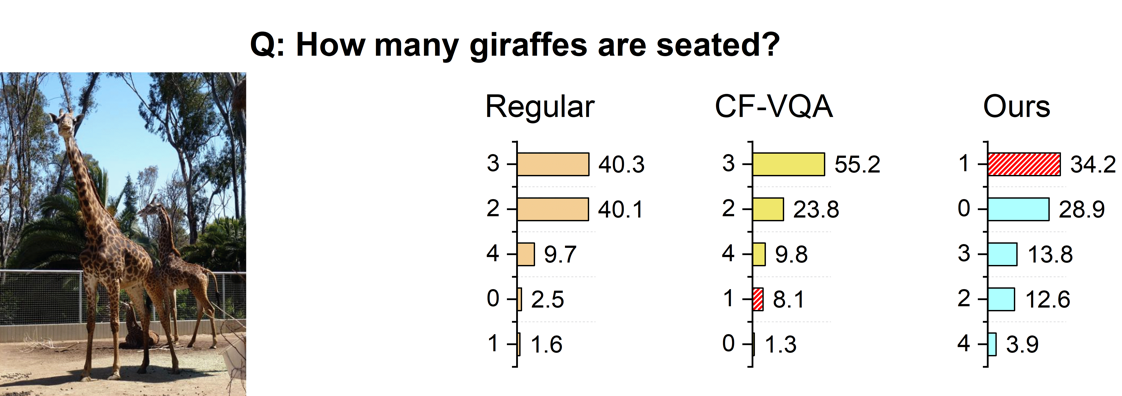}
     \end{subfigure}
     \hfill
     \begin{subfigure}[b]{0.49\textwidth}
         \centering
         \includegraphics[width=\textwidth]{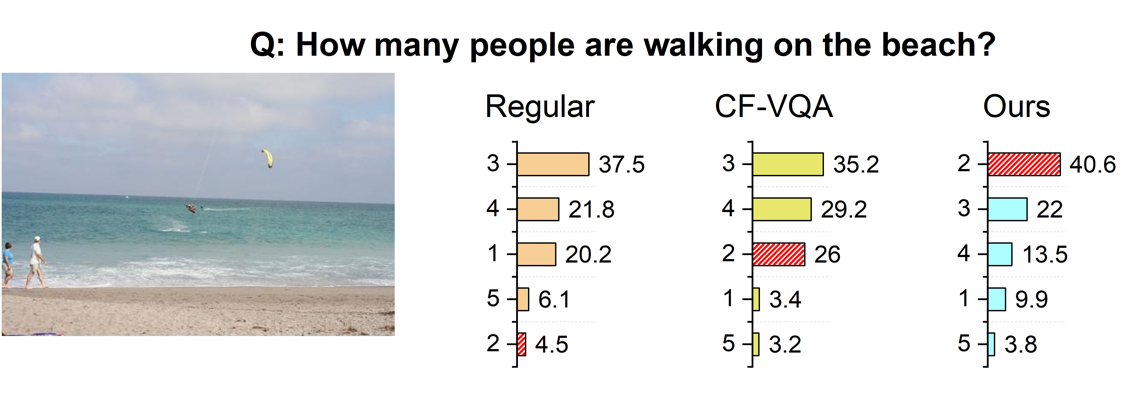}
     \end{subfigure}
     \hfill
     \begin{subfigure}[b]{0.49\textwidth}
         \centering
         \includegraphics[width=\textwidth]{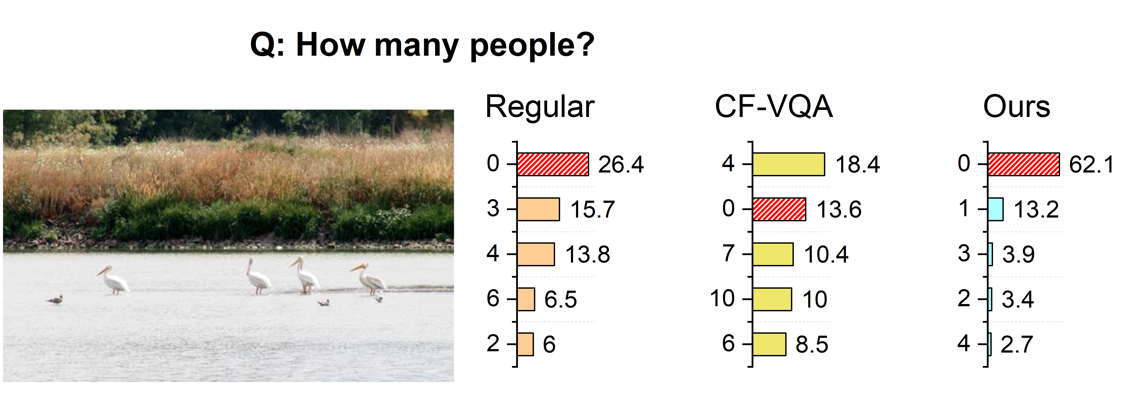}
     \end{subfigure}
     \hfill
     \begin{subfigure}[b]{0.49\textwidth}
         \centering
         \includegraphics[width=\textwidth]{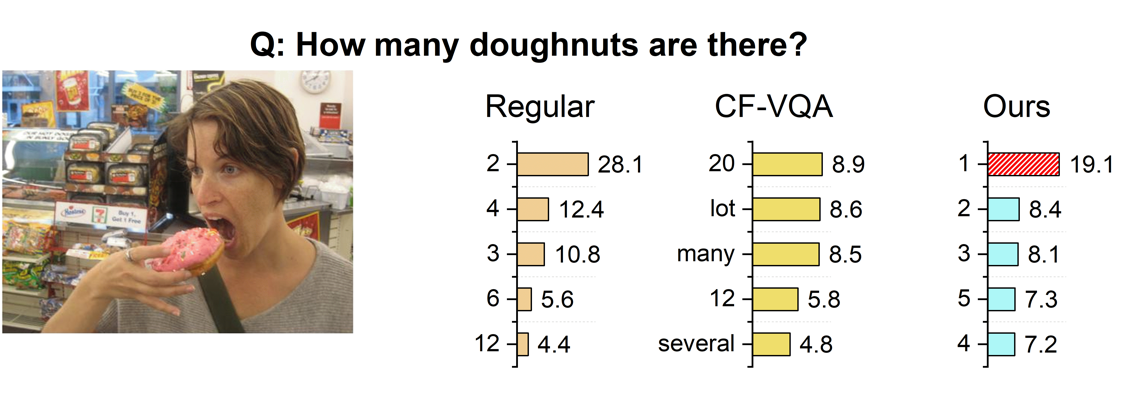}
     \end{subfigure}
     \hfill
     \begin{subfigure}[b]{0.49\textwidth}
         \centering
         \includegraphics[width=\textwidth]{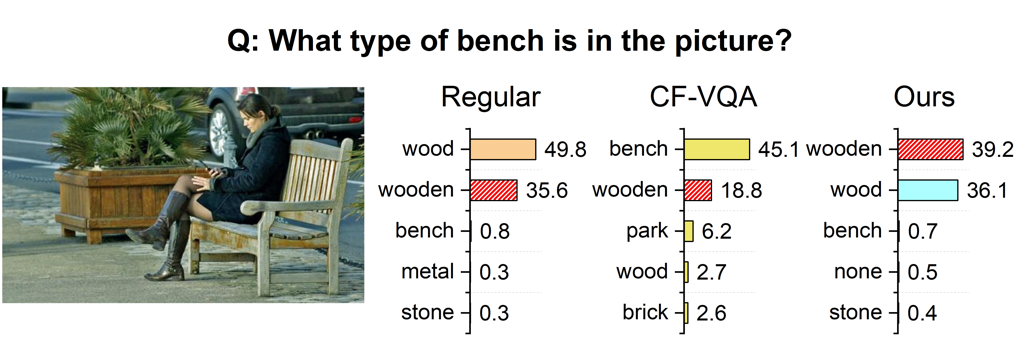}
     \end{subfigure}
        \caption{Qualitative comparison on VQA-CP v2 test split on regular VQA, CF-VQA~\cite{niu2021counterfactual} and our method are shown as bar plots, where the red bars with a sparse pattern are ground-truth. Values on the bars are probabilities out of 100\% to have an answer as correct.}
        \label{fig:qualitative_eavqa}
\end{figure*}

\begin{figure*}[h!t!b]
     \centering
     \begin{subfigure}[b]{0.31\textwidth}
         \centering
         \includegraphics[width=\textwidth]{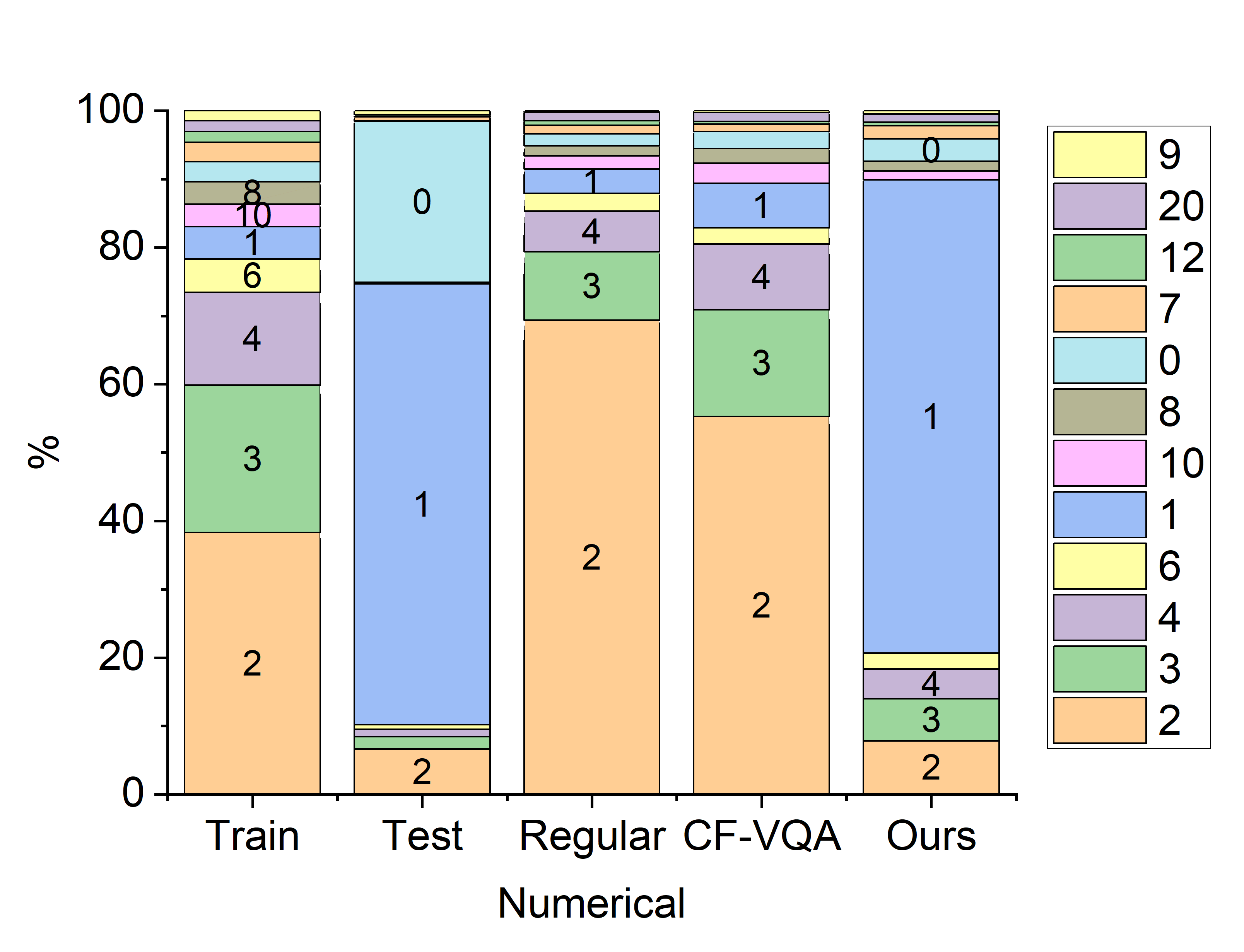}
     \end{subfigure}
     \hfill
     \begin{subfigure}[b]{0.31\textwidth}
         \centering
         \includegraphics[width=\textwidth]{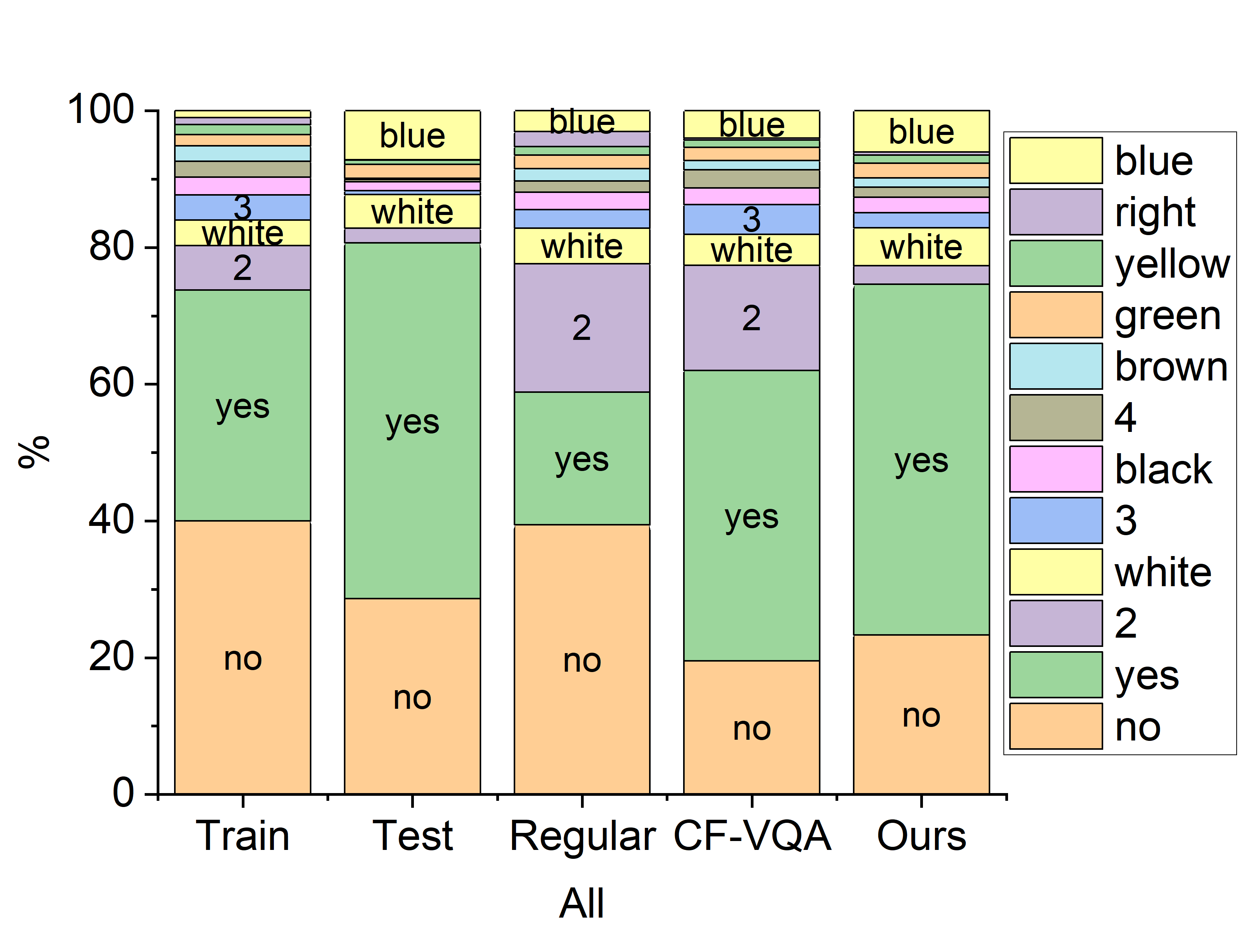}
     \end{subfigure}
     \hfill
     \begin{subfigure}[b]{0.31\textwidth}
         \centering
         \includegraphics[width=\textwidth]{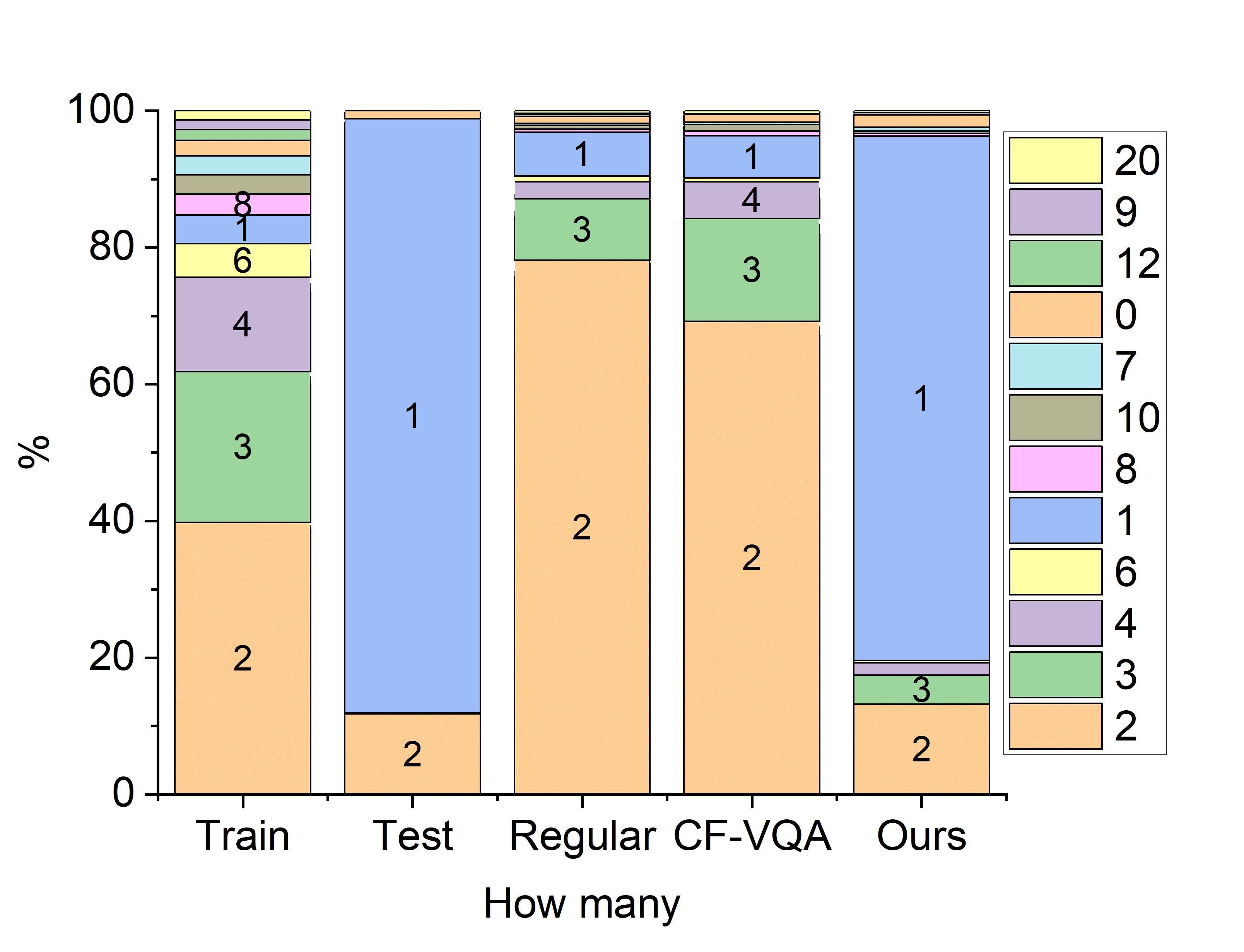}
     \end{subfigure}
        \caption{The distributions of the train, test sets, the previous methods, namely regular VQA, CF-VQA~\cite{niu2021counterfactual}, and our proposed method are shown. Note that there is a subtle difference between "how many" questions versus questions with "numerical" answers, which is related to the difference between recognizing "numerical" rather than counting "how many". }
        \label{fig:train_test_distributions}
\end{figure*}

As listed in Table~\ref{tab:all}, our method outperforms most of the competing methods on the benchmark datasets, especially in  numerical questions, which was introduced as an open problem recently~\cite{niu2021counterfactual}. Moreover, the results indicate that our method improves the accuracy of both the S-MRL and UpDn backbones, demonstrating that they are generalizable to both architectures. Noteworthy to mention that there are higher accuracy values for methods that augment data. In contrast, these methods are not comparable to ours as they do not obey the main idea of the VQA-CP v2 dataset, conducting unbiased inference under biased training. Simulation results of our proposed EA fusion strategy and the PW-VQA are shown in Fig.~\ref{fig:backbone_ablaton}. Both the EA fusion strategy and PW-VQA framework increase the accuracy of all question types. Particularly, the accuracy of numerical results with SAN as backbone increases from 12.4 to 37.6 by adding EA fusion and further increases to 57.7 by adding the PW-VQA framework. Furthermore, the improvements are consistent for all backbones, including SAN, S-MRL, and UpDn. More improvements can be achieved using large pretrained language-vision models. We used generative BLIP decoder~\cite{li2022blip} and CLIP  encoders~\cite{radford2021learning} to achieve more improvements.\footnote{For more explanations see the appendix.}\newline
\noindent\textbf{Qualitative results}
To qualitatively show the results of our method vs CF-VQA and regular VQA, we did simulations on the VQA-CP v2 dataset which, some examples have been shown in Fig.~\ref{fig:qualitative_eavqa}. As seen in the pictures of Fig.~\ref{fig:qualitative_eavqa}, our method is less biased by either language or vision bias. For example, When asked, "Is she eating?" for a picture showing a cat and a woman, our method can correctly answer "No" while the regular VQA cannot. Interestingly, the CF-VQA is clearly biased by the salient object in the picture and has extremely high confidence in answering this question with "cat" which is obviously ridiculous.
Another example is both regular VQA and CF-VQA cannot use the key information from the question, which results in answering the question with biased inference due to the foreground animals in the picture. More examples are in the appendix. 

Five distributions of numerical answers for train, test, regular VQA, CF-VQA, and our model are shown in Fig.~\ref{fig:train_test_distributions}. We can clearly see in Fig.~\ref{fig:train_test_distributions} that CF-VQA captured many biases in question with "numerical" answers from the training dataset. At the same time, our model reduces those biases and obtains answer distribution benefits from removing language and vision biases simultaneously during de-confounding causal inference and closer to the test dataset.

\section{Related Work}\label{sec:relatedWorks}
VQA-CP dataset has been proposed to benchmark the generalizability of VQA models under changing prior conditions \cite{agrawal2018don}. Various methods \cite{niu2021counterfactual, han2021greedy, gat2020removing, gouthaman2020reducing, abbasnejad2020counterfactual, kolling2022efficient, gupta2022swapmix, shrestha2022investigation} have been proposed to solve VQA-CP, which can be divided into four main categories. 1) Methods that modify language module or use language prior to suppressing or controlling language shortcuts by separating question-only branches or capturing language prior to subtracting or masking in the model~\cite{ramakrishnan2018overcoming,cadene2019rubi,clark2019don}. 2) Methods that mitigate bias through reducing visual bias or enhancing visual attention/grounding use human input to increase the attention to visual information or reduce contextual biases that shortcut vision to answer~\cite{selvaraju2019taking,wu2019self,das2017human}. 3) Mitigation of both language and vision bias together that tries to balance two modalities of vision and language for robust multimodal inference. And 4) Methods that synthesize data to augment and balance training splits, that use generative models to synthesize and augment visual and linguistic data to balance the distribution of training splits~\cite{chen2020counterfactual,abbasnejad2020counterfactual,teney2020unshuffling,zhu2020overcoming, kolling2022efficient}.

The causal inference has inspired several studies in computer vision, including visual explanations~\cite{goyal2019counterfactual,wang2020scout,yi2019clevrer}, scene graph generation~\cite{tang2020unbiased}, image recognition~\cite{tang2020unbiased}, zero-shot and few-shot learning~\cite{yue2020interventional,yue2021counterfactual} incremental learning~\cite{hu2021distilling}, representation learning~\cite{wang2020visual,zhang2020devlbert}, semantic segmentation~\cite{zhang2020causal}, and vision-language tasks~\cite{chen2020counterfactual,teney2020learning,yang2020deconfounded,fu2020sscr,yang2021causal}. 
Especially, counterfactual learning has been exploited in recent VQA studies~\cite{chen2020counterfactual,teney2020learning,abbasnejad2020counterfactual}. 

\section{Conclusion} \label{sec:conclusion}
VQA systems suffer from leveraging information only from one modality, especially the language modality from the given question. Many methods have been proposed to address this kind of problem. However, the previous method didn't consider that biases that come from each modality are highly confounded through the annotation process. VQA systems that ignore this effect cannot avoid increasing the bias learned from one modality while trying to reduce bias from another modality. We formulate the Explain-Away effect that causes the bias of both vision and language modalities with a novel causal framework for VQA systems. This framework can be implemented on the different VQA backbones and improve their generalizability significantly. The proposed framework successfully helps VQA systems reduce language bias without increasing vision bias. Experiment results show that our proposed method achieved state-of-the-art performance on de-bias oriented dataset VQA-CP especially doubled the accuracy on numerical questions from the previous best model.
\section*{Acknowledgments}
This work has been partially supported by the National Science Foundation (NSF) under Grant 1909912 and the Defense Advance Research Projects Agency (DARPA) under Contract HR00112220003. The content of the information does not necessarily reflect the position of the Government, and no official endorsement should be inferred.

\bibliographystyle{unsrtnat}
\bibliography{references}  






\appendix

\section*{Limitations}
\begin{figure}[h!t!b]
     \centering
     \begin{subfigure}[b]{0.49\textwidth}
         \centering
         \includegraphics[width=\textwidth]{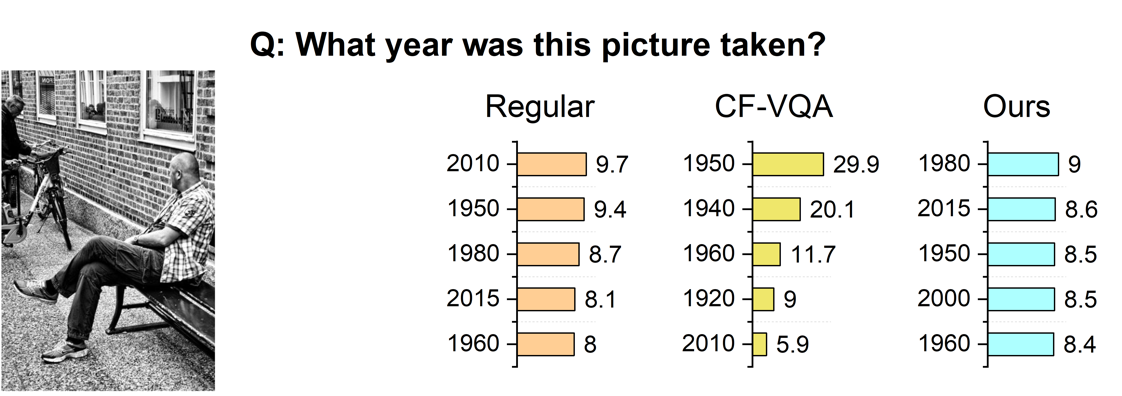}
         \label{fig:five over x}
     \end{subfigure}
     \hfill
     \begin{subfigure}[b]{0.49\textwidth}
         \centering
         \includegraphics[width=\textwidth]{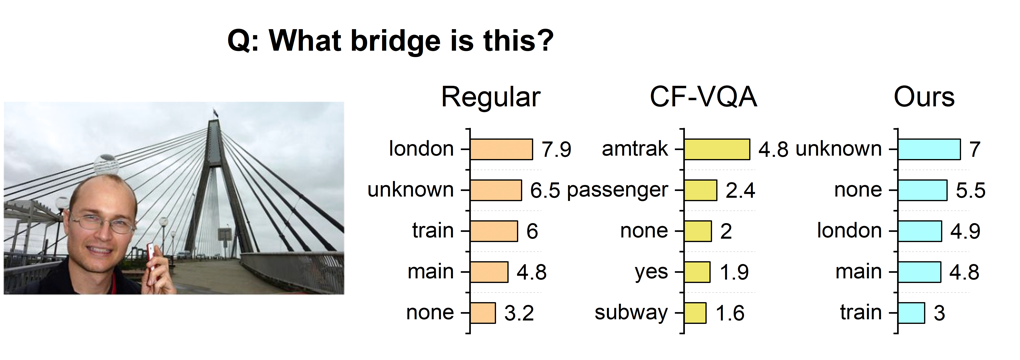}
         \label{fig:five over x}
     \end{subfigure}
        \caption{A limitation of VQA models is shown with two examples where PW-VQA fails to answer correctly; however, CF-VQA and regular VQA also fail.}
        \label{fig:limitations}
\end{figure}

For questions and images that require background knowledge, PW-VQA fails to answer correctly. Indeed, CF-VQA and regular VQA also fail, as shown in Fig. \ref{fig:limitations}. For instance, the answer to the question "What year was this picture taken?" requires background knowledge of details such as the years that those sneakers and bicycles were used. Given this knowledge, the answer may be accurate within a range, for instance, 1960-1980, which requires open-ended answers with visual reasoning of small and large details in both image and language. Likewise, the "what bridge is this?" answer requires background knowledge. Overall, our method and all other VQA systems require background knowledge to answer some questions.

\section*{Ethics Statement}
VQA systems are fundamental building blocks in many AI systems, including visual dialog and question-answering systems. Therefore, any unethical information from the system, such as racial or gender-related problems, may have adverse social impacts if used at scale. Overall, VQA systems are helpful as AI assistants can aid people with disability or visual impairments or be used for visual queries with natural language interfaces.

\section{Appendix}

\subsection{Architecture of the PW-VQA}
\begin{figure*}[t]
    \centering
    \includegraphics[width=0.7\linewidth]{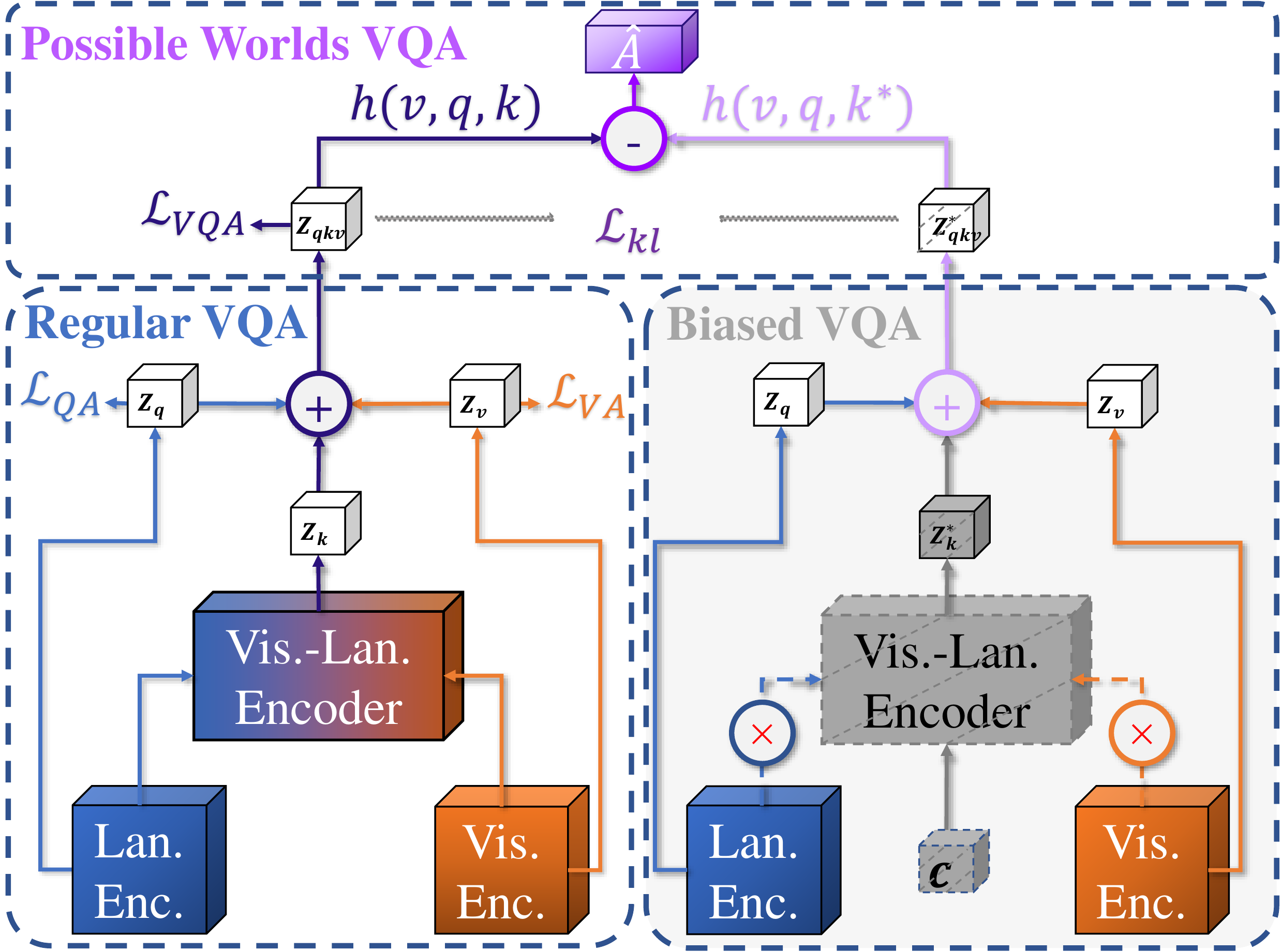}
    \caption{The image demonstrates our proposed PW-VQA framework.     }
    \label{fig:neural_network_pwvqa}
\end{figure*}

For the page limit considerations, we moved the PW-VQA framework's architecture to here, shown in Fig.~\ref{fig:neural_network_pwvqa}. The framework consists of two branches: regular VQA, which can be of any baseline method, and the counterfactual version of the same network and parameters, shown as a biased branch in the figure. Four different losses are simultaneously used during training to formulate the causal relationship between each modality. In addition, a constant $c$ is jointly trained here to capture the natural indirect effect of vision-language confounding biased injected during the annotation process. Finally, in the inference stage, PW-VQA uses the logits of regular VQA subtracted by the biased VQA and gets a debiased answer. The letters $a$ in this figure denote the answer. 

\subsection{Details on CLIP-BLIP model as baseline and results}
\begin{figure*}[t]
    \centering
    \includegraphics[width=0.8\linewidth]{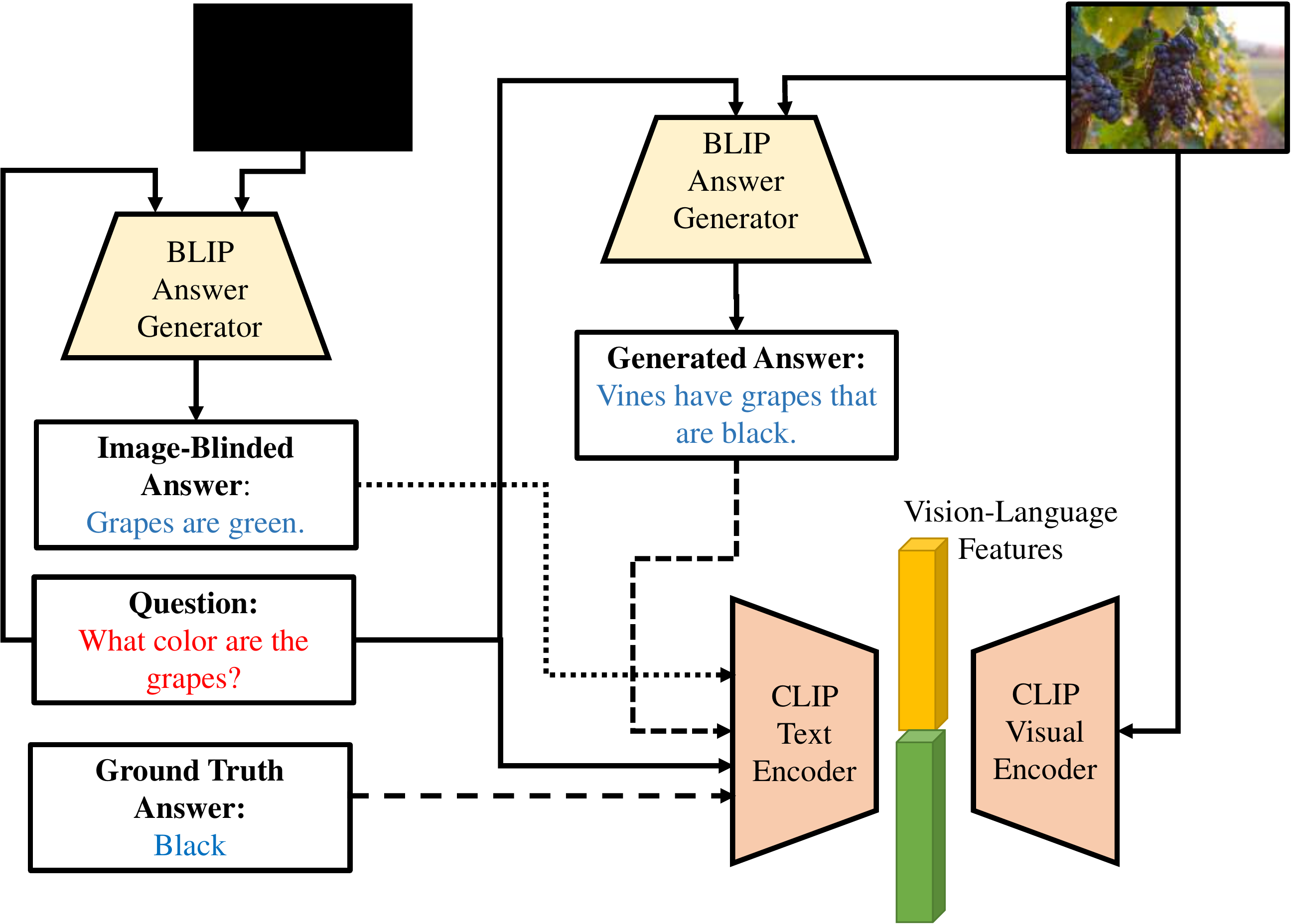}
    \caption{The architecture of CLIP-BLIP network as a baseline for PW-VQA. We combine generative pretrained BLIP \cite{li2022blip} with encoding of the CLIP \cite{radford2021learning} as a transformer-based large vision-language model.}
    \label{fig:clip_blip_architecture}
\end{figure*}

Contrastive Language-Image Pretraining (CLIP) is a model trained on large image-text pairs with a contrasting loss \cite{radford2021learning}. We use CLIP as the baseline for implementation as an ablation study on large pretrained vision-language models. Our implementation uses CLIP as a text and image encoder without fine-tuning. During the training, we use $(v,q, a)$ ground-truth tuples and tokenize both question and answer. Then use CLIP to encode text and question tokens and normalize their sum. Then we use the CLIP image encoder and normalize features to embed images for the classification. We also use MLP layers to obtain logits of vision-only and language-only branches to later use in PW-VQA. The fusion of image and language is based on concatenation, and an MLP is used to obtain output vision-language logits.
During inference, we use Bootstrapping Language-Image Pre-training for Unified Vision-Language Understanding and Generation (BLIP) \cite{li2022blip}, a generative pretrained model for open-ended visual question answering, and then use CLIP tokenizer and encoder to obtain language-only and vision-only and language-vision logits to be used in PW-VQA. As BLIP uses questions and images to generate open-ended answers, we use zero tensors as image inputs to the BLIP model to avoid spilling image features to question-only logits. Finally, CLIP features are normalized and used in the framework. Results for the BLIP-CLIP model as the baseline and other models are listed in Tab. \ref{tab:backbone_effect_clip_blip}. Although the accuracy of the BLIP-CLIP is significantly higher when used as a baseline, it cannot be compared with the other baselines, as BLIP is pretrained on more than 400 million image-text pairs, some of which are high-quality data. Similarly, CLIP is pretrained on large datasets of image-text pairs that have high-quality image-text pairs. Figure \ref{fig:clip_blip_architecture} shows the architecture of the implemented CLIP-BLIP model.

\begin{table*}[!t]
\centering
\caption{The table lists the accuracy values for different backbones based on VQA-CP v2 and VQA v2 datasets. We use different backbones, UpDn, S-MRL, and CLIP-BLIP, to show the effect of the backbone on the accuracy. We show the best-performing method with bold and the second-best-performing method with an underline.}
\label{tab:backbone_effect_clip_blip}
\scalebox{1.0}{
\begin{tabular}{|llrrrrrrrr|}
\hline
Test set                                                         &       & \multicolumn{4}{|l|}{VQA-CP v2 test}                                                                       & \multicolumn{4}{|l|}{VQA v2 test}                                                                         \\ \cline{3-10} 
Methods                                                          & Base  & \multicolumn{1}{|l|}{All} & \multicolumn{1}{|l|}{Y/N} & \multicolumn{1}{|l|}{Num.} & \multicolumn{1}{|l|}{Other} & \multicolumn{1}{|l|}{All} & \multicolumn{1}{|l|}{Y/N} & \multicolumn{1}{|l|}{Num.} & \multicolumn{1}{|l|}{Other} \\ \hline
PW-VQA (ours)                                                     & UpDn  & {59.06}                   & \underline{88.26}                   & {52.89}                    & 45.45                     & \underline{62.63}                   & \underline{81.80}                   & \underline{43.90}                    & \underline{53.01}                     \\
PW-VQA (ours)                                                     & S-MRL & \underline{60.26}                   & 88.09                   & \underline{59.13}                    & \underline{45.99}                     & 61.25                   & 80.32                   & 43.17                    & 51.53                     \\ 
PW-VQA (ours)                                                     & CLIP-BLIP & \textbf{76.57}                   & \textbf{97.23}                   & \textbf{69.39}                    & \textbf{67.72}                     & \textbf{78.17}                   & \textbf{97.27}                   & \textbf{62.26}                    & \textbf{67.85}                     \\ \hline
\end{tabular}
}
\end{table*}

\subsection{More Qualitative Examples}
\begin{figure*}[h!t!b]
     \centering
     \begin{subfigure}[b]{0.49\textwidth}
         \centering
         \includegraphics[width=\textwidth]{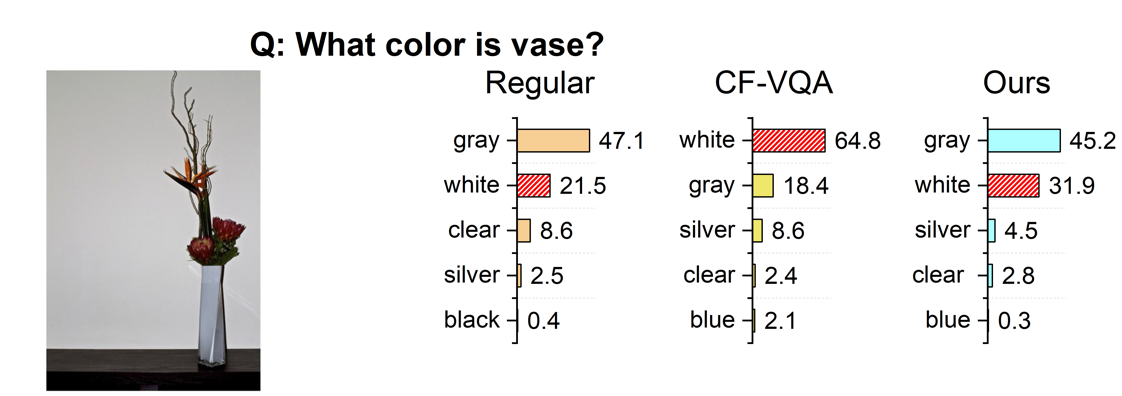}
     \end{subfigure}
     \hfill
     \begin{subfigure}[b]{0.49\textwidth}
         \centering
         \includegraphics[width=\textwidth]{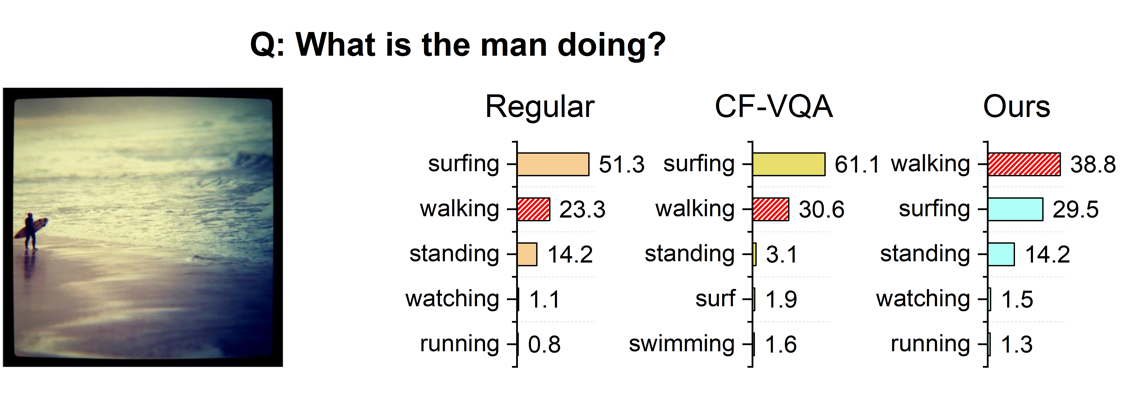}
     \end{subfigure}
     \hfill
     \begin{subfigure}[b]{0.49\textwidth}
         \centering
         \includegraphics[width=\textwidth]{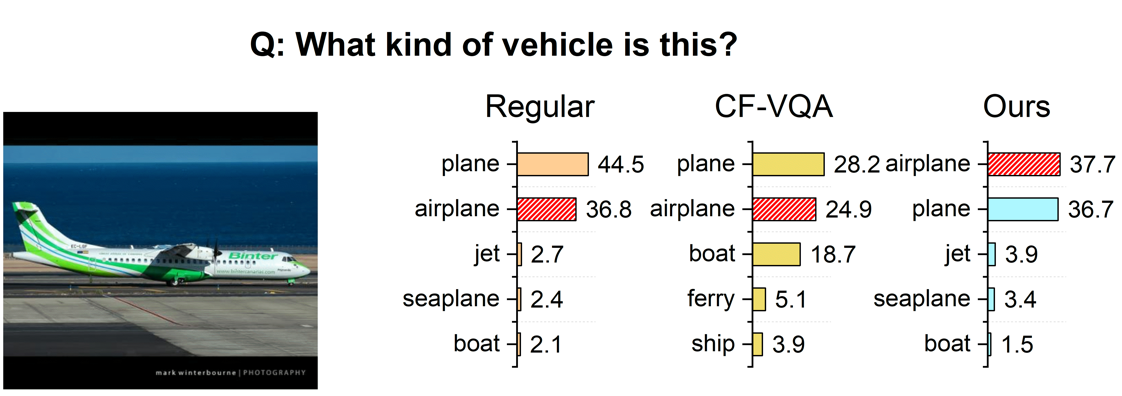}
     \end{subfigure}
     \hfill
     \begin{subfigure}[b]{0.49\textwidth}
         \centering
         \includegraphics[width=\textwidth]{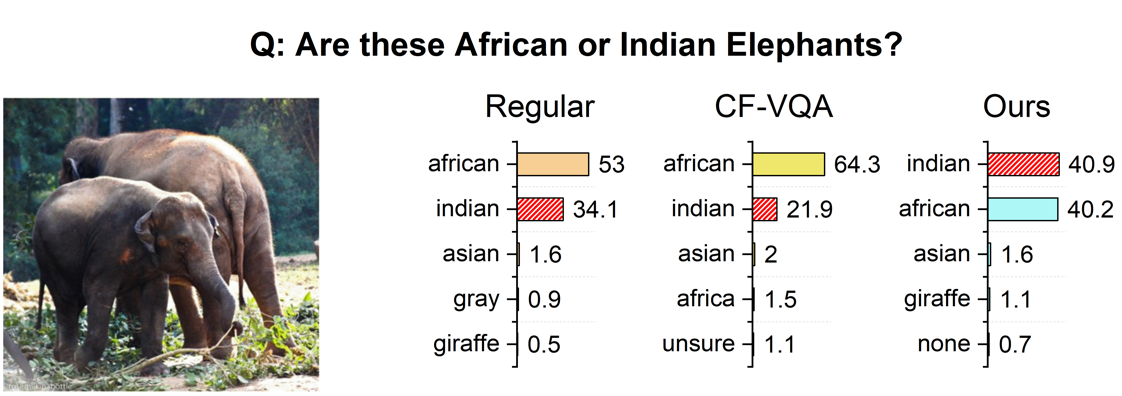}
     \end{subfigure}
     \hfill
     \begin{subfigure}[b]{0.49\textwidth}
         \centering
         \includegraphics[width=\textwidth]{figures/Qual_example_195264.png}
     \end{subfigure}
     \hfill
     \begin{subfigure}[b]{0.49\textwidth}
         \centering
         \includegraphics[width=\textwidth]{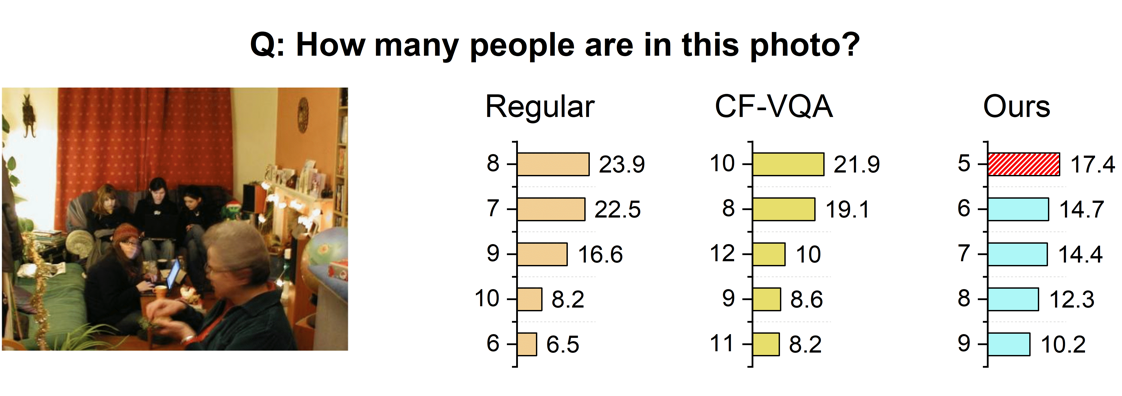}
     \end{subfigure}
     \hfill
     \begin{subfigure}[b]{0.49\textwidth}
         \centering
         \includegraphics[width=\textwidth]{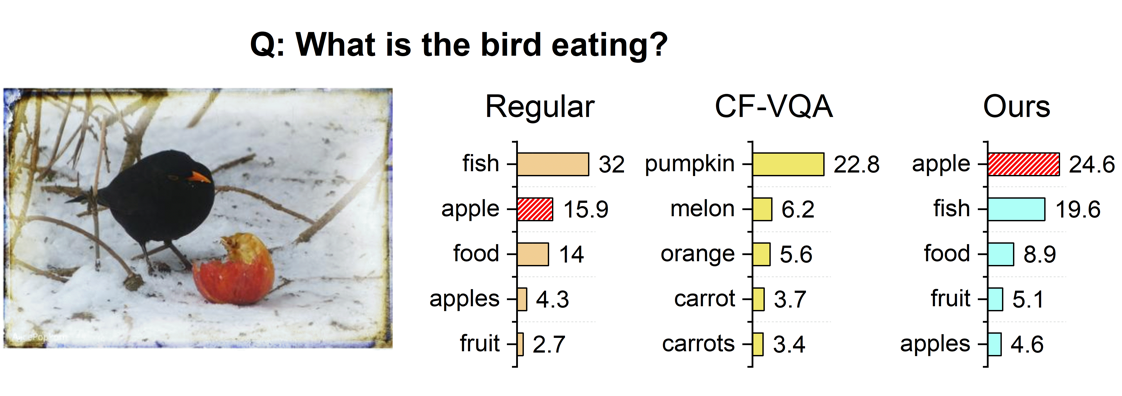}
     \end{subfigure}
     \hfill
     \begin{subfigure}[b]{0.49\textwidth}
         \centering
         \includegraphics[width=\textwidth]{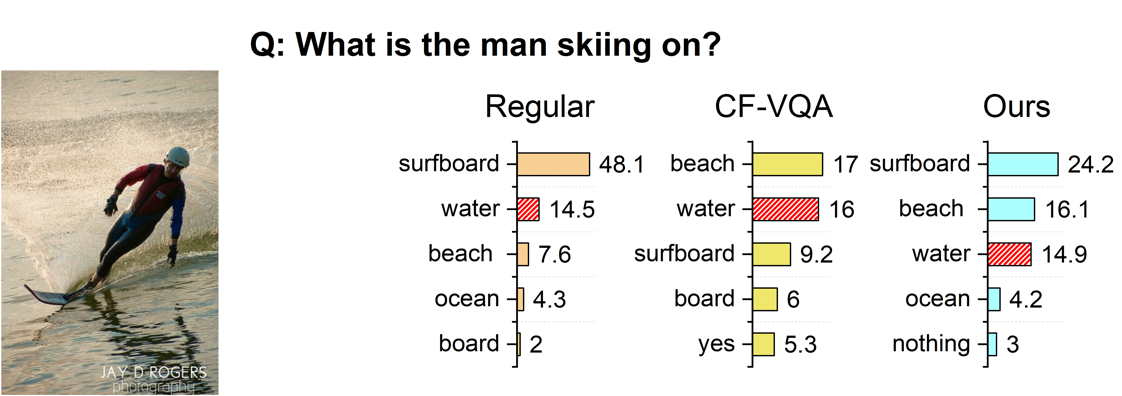}
     \end{subfigure}
     \hfill
     \begin{subfigure}[b]{0.49\textwidth}
         \centering
         \includegraphics[width=\textwidth]{figures/Qual_example_195447.png}
     \end{subfigure}
     \hfill
     \begin{subfigure}[b]{0.49\textwidth}
         \centering
         \includegraphics[width=\textwidth]{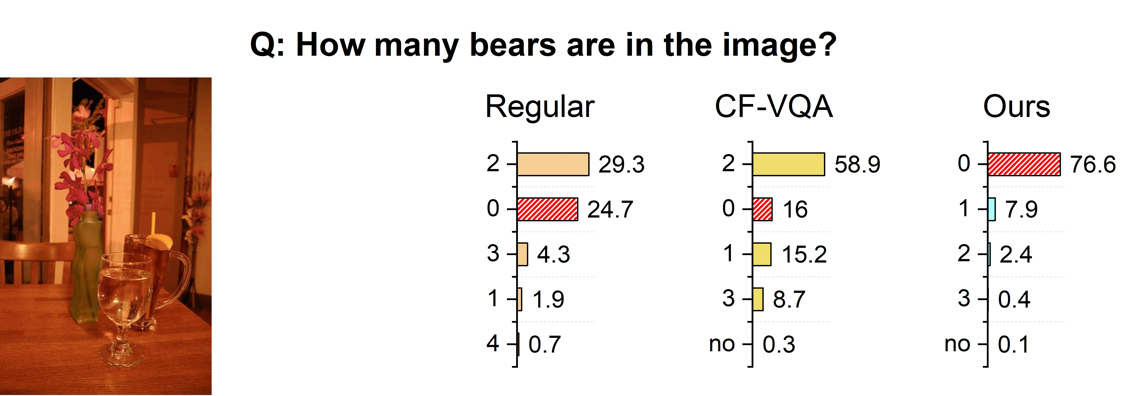}
     \end{subfigure}
     \hfill
     \begin{subfigure}[b]{0.49\textwidth}
         \centering
         \includegraphics[width=\textwidth]{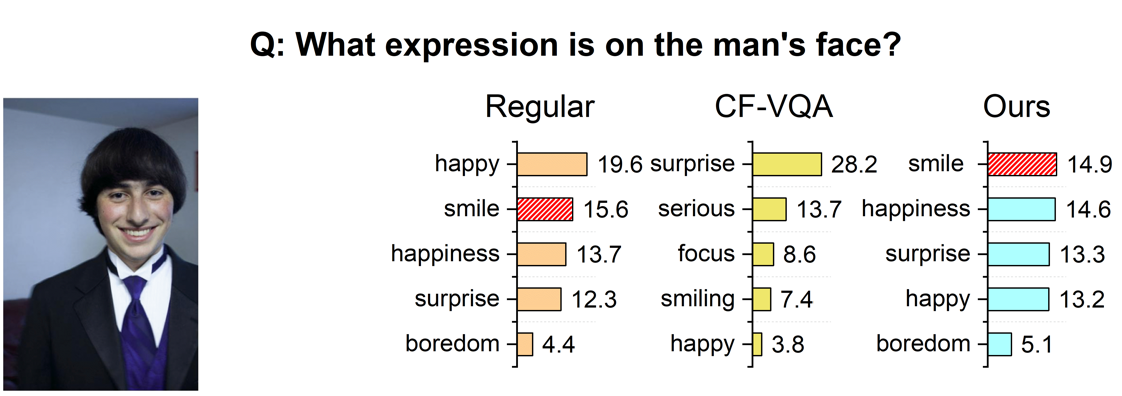}
     \end{subfigure}
     \hfill
     \begin{subfigure}[b]{0.49\textwidth}
         \centering
         \includegraphics[width=\textwidth]{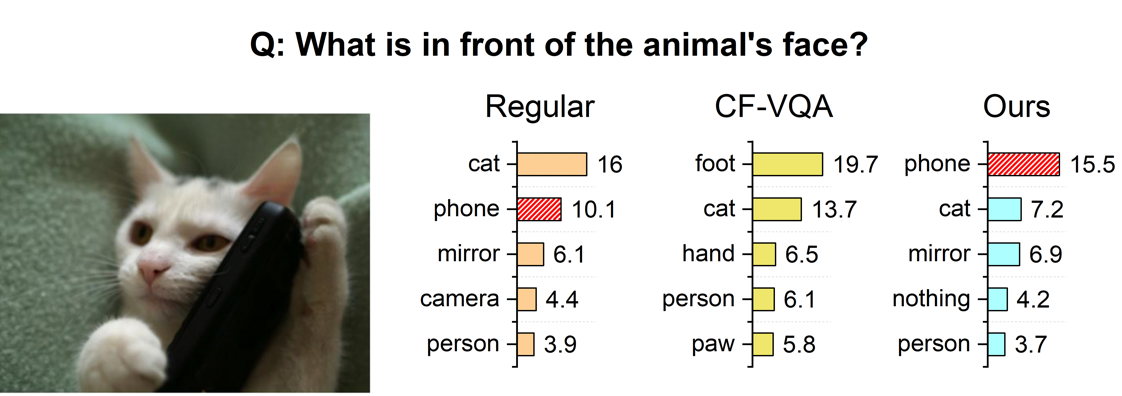}
     \end{subfigure}
        \caption{Qualitative comparison of VQA-CP v2 test split, our method vs. CF-VQA \citep{niu2021counterfactual} and regular VQA are shown in these images. Red bars denote the ground-truth one, while the other bars denote the prediction probability corresponding to their value.}
        \label{fig:qualitative_examples_appendix}
\end{figure*}

Qualitative comparison of VQA-CP v2 test split, our method vs. CF-VQA \citep{niu2021counterfactual} and regular VQA are shown in Fig. \ref{fig:qualitative_examples_appendix}. Red bars denote the ground-truth one, while the other bars denote the prediction probability corresponding to their value. As shown in this figure, all methods fail on the question and image pairs that require background knowledge and reasoning.

\subsection{Stabilizing logarithmic computations}
\begin{figure*}
    \centering
    \includegraphics[width=0.5\linewidth]{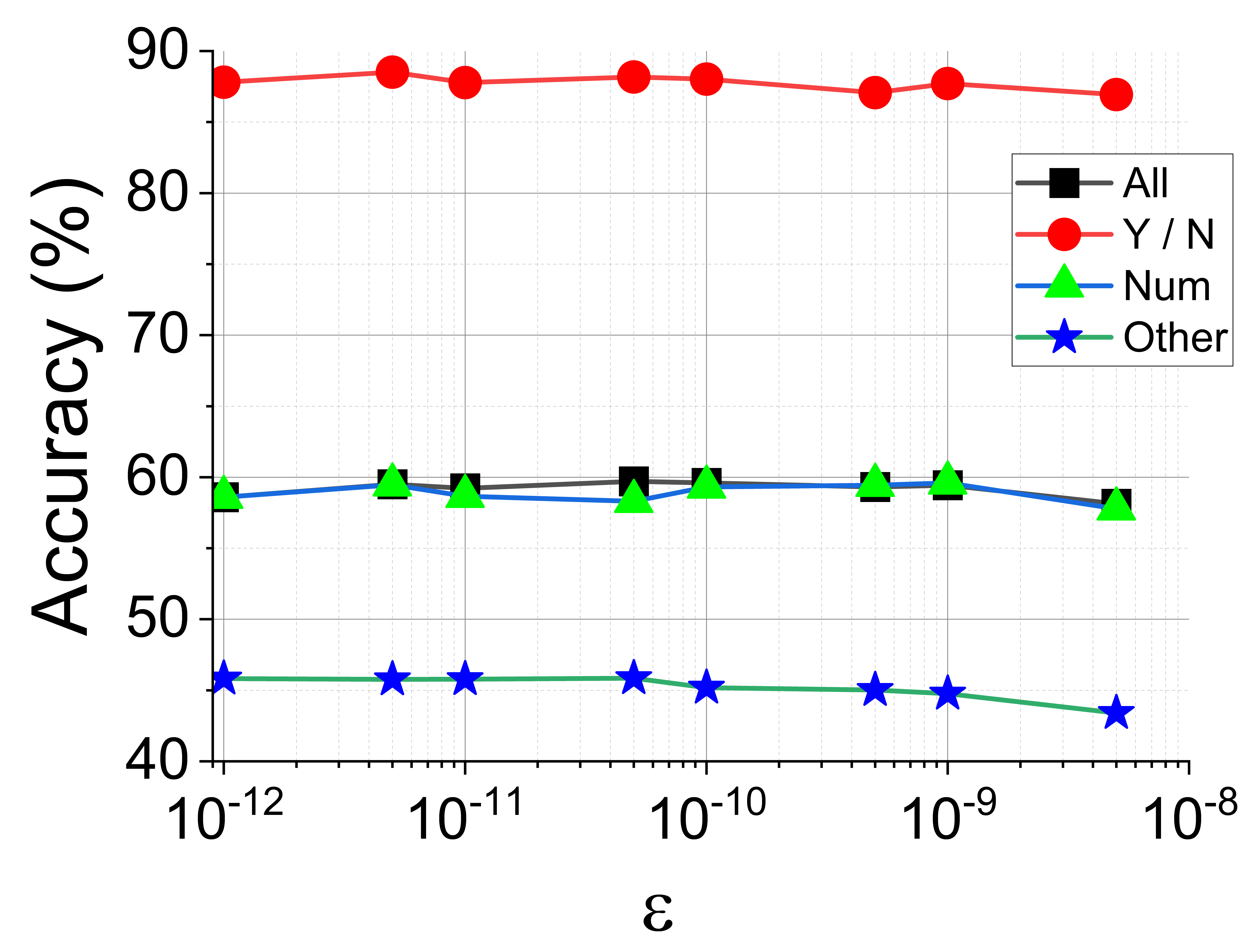}
    \caption{Ablation of different values of epsilon on VQA-CP v2 test set. Variations of $\epsilon$ has a slight effect on improving the results, though the reason may be related to computational stability. These results are related to PW-VQA with $\alpha = 1.5$ and $\epsilon = \{ 10^{-12} , 5 \times 10^{-12} , \ldots , 5 \times 10^{-9} \}$.}
    \label{fig:Epsilon_sweep}
\end{figure*}

We noticed that adding a constant $\epsilon$ to the logarithmic computations of the fusion equations is important for the training and test consistency. For this reason, we also tried different $\epsilon$ to stabilize logarithmic computations of the loss function and include it in the ablation study. The results are reported in Tab. ~\ref{tab:epsilon}, and are shown in Fig. ~\ref{fig:Epsilon_sweep}.

\begin{table}[t]
\centering
\caption{Ablation of different values of $\epsilon$ on VQA-CP v2 test set. The backbone here is the SMRL network. }
\label{tab:epsilon}
\scalebox{1}{
\begin{tabular}{r|rrrr}
\hline
\multicolumn{1}{l}{$\epsilon$} & \multicolumn{1}{l}{All} & \multicolumn{1}{l}{Y / N} & \multicolumn{1}{l}{Num} & \multicolumn{1}{l}{Other} \\ \hline
1.00E-12                    & 58.6                    & 87.81                     & 58.59                   & 45.82                     \\ \hline
5.00E-12                    & 59.51                   & 88.51                     & 59.47                   & 45.77                     \\ \hline
1.00E-11                    & 59.22                   & 87.78                     & 58.66                   & 45.79                     \\ \hline
5.00E-11                    & 59.71                   & 88.18                     & 58.31                   & 45.85                     \\ \hline
1.00E-10                    & 59.6                    & 88.03                     & 59.32                   & 45.19                     \\ \hline
5.00E-10                    & 59.31                   & 87.07                     & 59.44                   & 45.02                     \\ \hline
1.00E-09                    & 59.44                   & 87.71                     & 59.6                    & 44.77                     \\ \hline
5.00E-09                    & 58.13                   & 86.94                     & 57.78                   & 43.41                     \\ \hline
\end{tabular}
}
\end{table}[t]

\subsection{Categorized improvements on SAN, UpDn, and SMRL baselines}
\begin{figure*}
     \centering
     \begin{subfigure}[b]{0.31\textwidth}
         \centering
         \includegraphics[width=\textwidth]{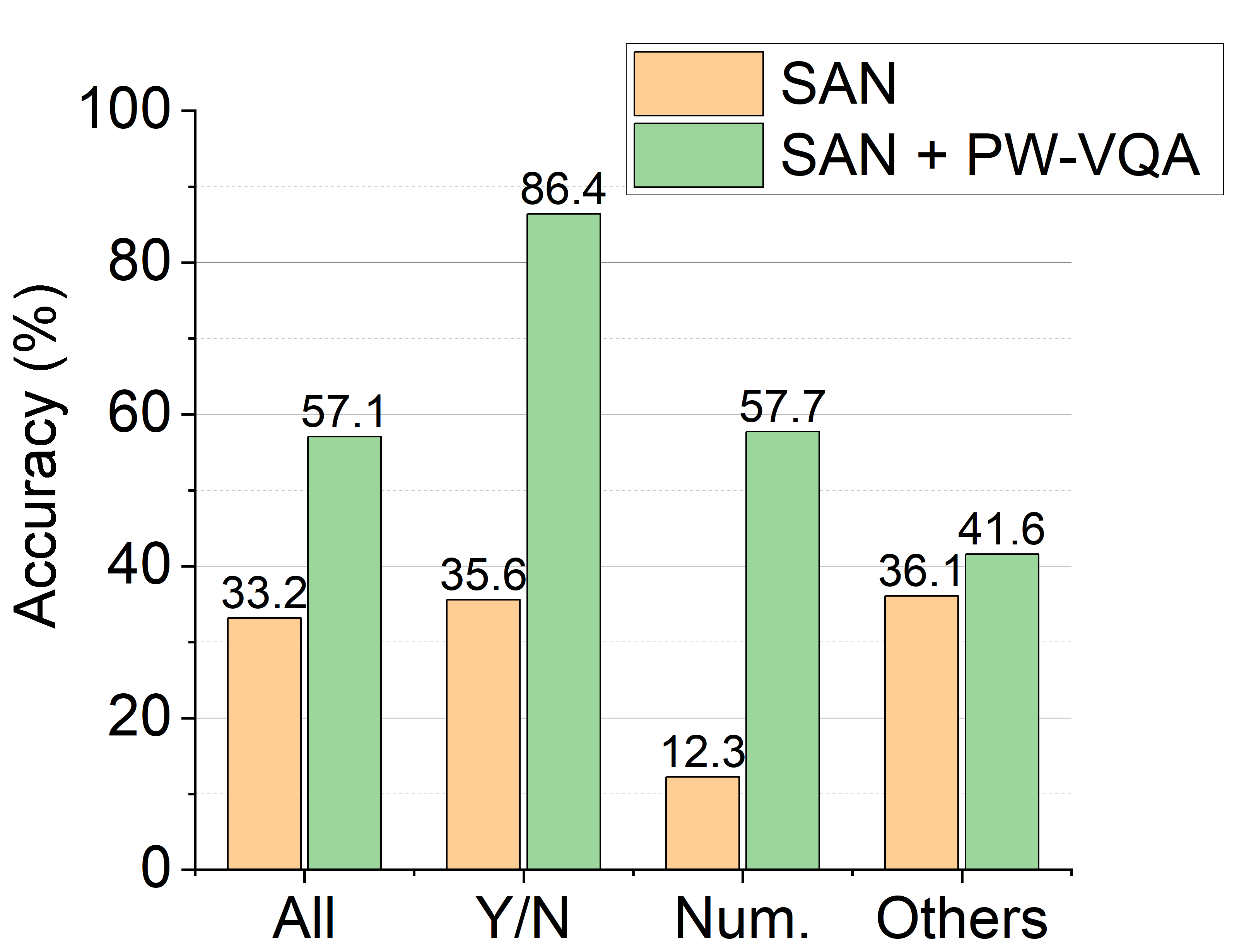}
     \end{subfigure}
     \hfill
     \begin{subfigure}[b]{0.31\textwidth}
         \centering
         \includegraphics[width=\textwidth]{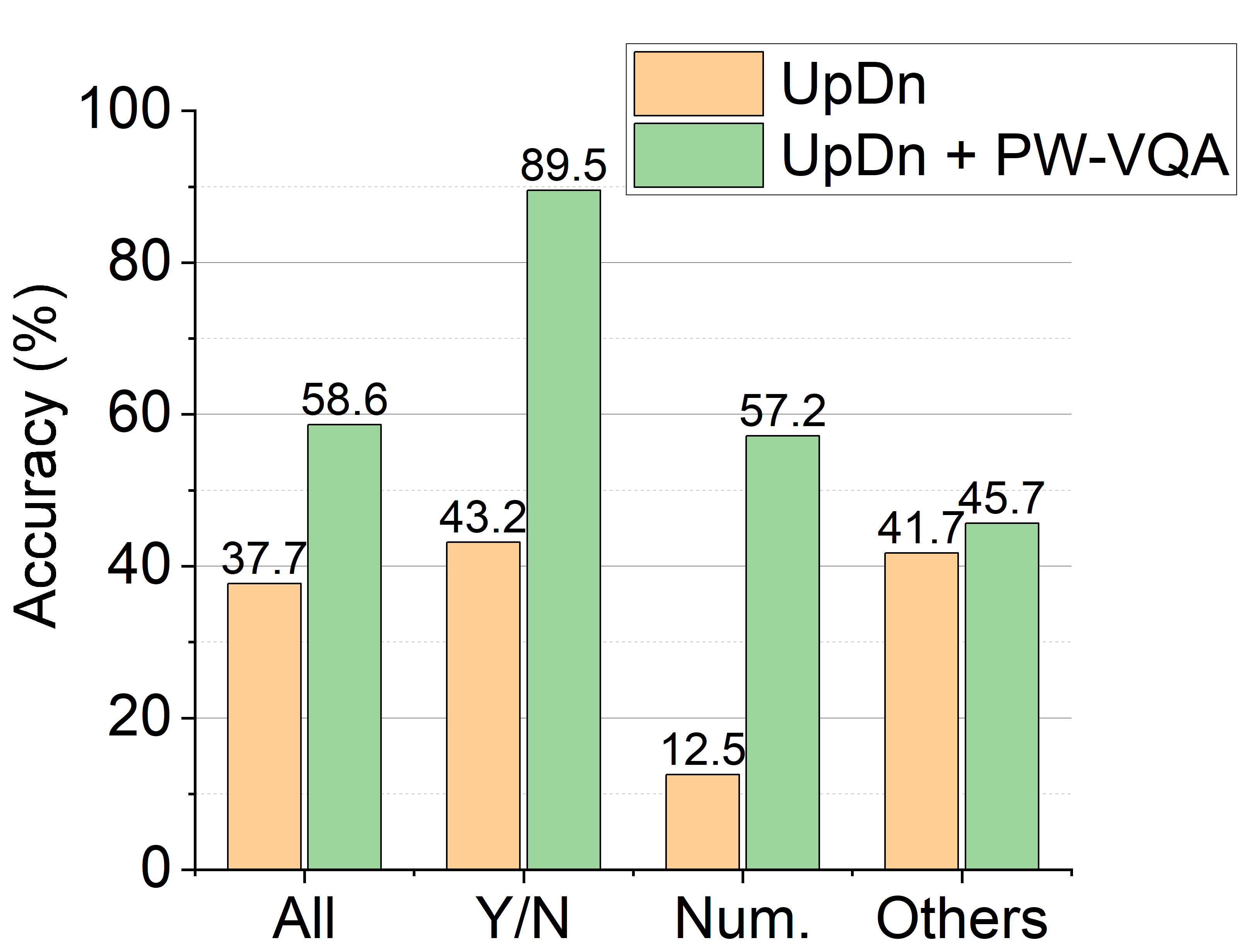}
     \end{subfigure}
     \hfill
     \begin{subfigure}[b]{0.31\textwidth}
         \centering
         \includegraphics[width=\textwidth]{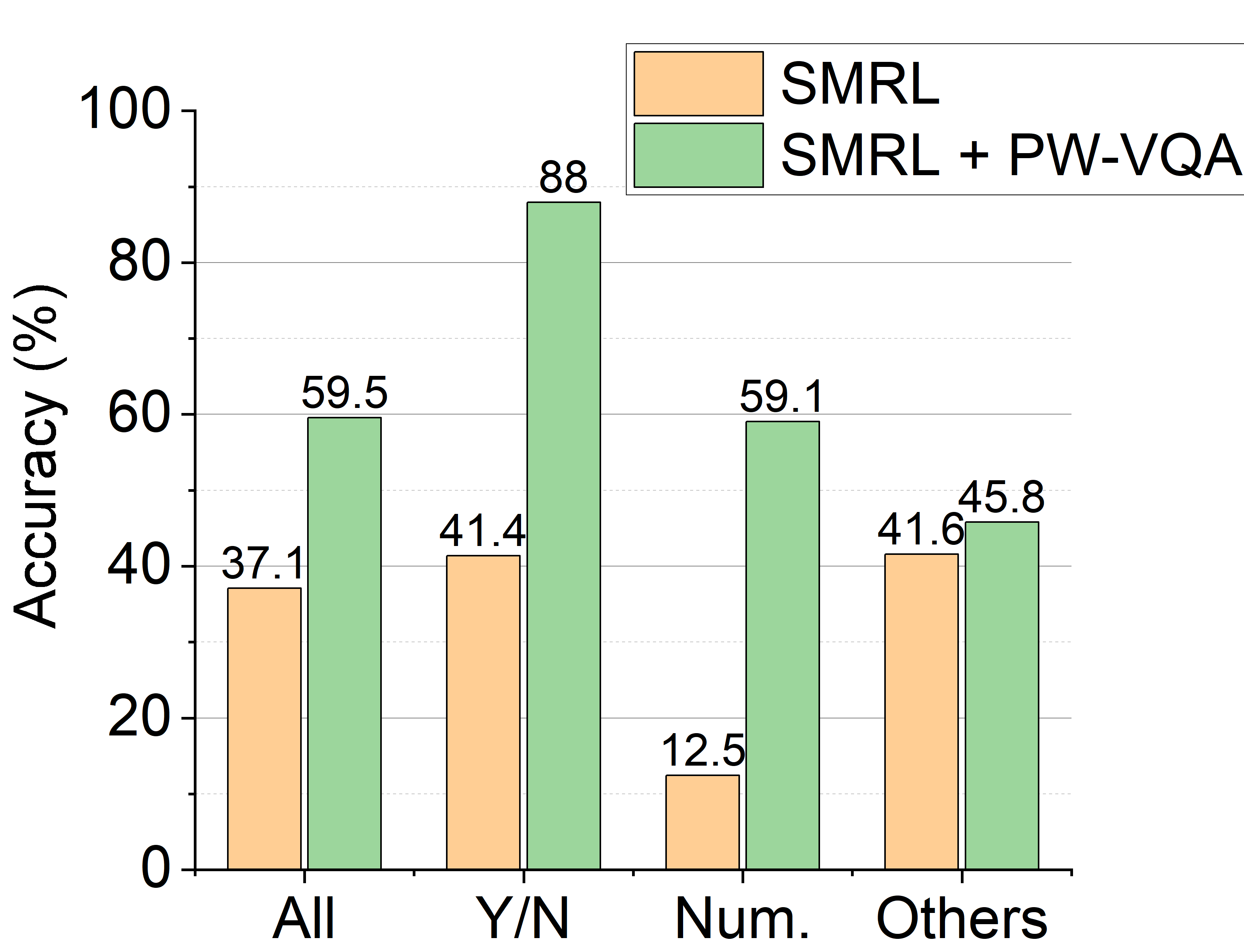}
     \end{subfigure}
        \caption{The plots here show the performance metrics in percent for different backbones using our proposed method, PW-VQA. All $\alpha = 1.4$ are set for all these simulations. Digits on the bars are rounded up to one digit. As shown here, the proposed method consistently improves the performance of all the backbone.}
        \label{fig:categorized_improvements}
\end{figure*}

The plots in Fig. \ref{fig:categorized_improvements} show performance metrics for different methods as baseline and percent of improvements compared to baseline on each class of question types when using our proposed method, PW-VQA. In all of these simulations, $\alpha = 1.4$ are set. As seen in Fig. \ref{fig:categorized_improvements}, PW-VQA consistently improves the existing method, confirming the generalizability of the method to several existing methods.

\subsection{Statistical analysis of performance values}
Variances of the accuracy results and mean values of the method are listed in Tab. \ref{tab:variances}.
\begin{table*}[!h]
\centering
\caption{Variances of the accuracy performance of our method based on five performing simulations with different random seeds.}
\label{tab:variances}
\scalebox{1}{

\begin{tabular}{|l|l|llll|}
\hline
Dataset   & Base  & Overall &  Y/N & Num. & Other \\ \hline
VQA v2    & S-MRL & 60.76$^{\pm 0.10}$         & 79.60$^{\pm 0.47}$          & 42.75$^{\pm 0.07}$        & 51.21$^{\pm 0.03}$        \\ \hline
VQA v2    & UpDn  & 62.62$^{\pm 0.01}$         & 81.80$^{\pm 0.06}$         & 43.59$^{\pm 0.05}$        & 53.09$^{\pm 0.01}$        \\ \hline
VQA-CP v2 & S-MRL & 59.71$^{\pm 0.14}$         & 87.98$^{\pm 0.24}$         & 59.41$^{\pm 0.61}$        & 45.51$^{\pm 0.09}$        \\ \hline
VQA-CP v2 & UpDn  & 58.70$^{\pm 0.20}$          & 89.19$^{\pm 0.23}$         & 58.85$^{\pm 0.56}$        & 45.17$^{\pm 0.05}$        \\ \hline
\end{tabular}

}
\end{table*}

\subsection{Causal graph improvements}
Table \ref{tab:ablation_backbones} presents an ablation study on the effect of the backbone and how our causal graph and proposed method improve the results. These results are excluding the EA fusion strategy and therefore, are to study the effect of a solely causal counterfactual mechanism that is to block vision-fusion collider bias. 
\begin{table*}[t]
\centering
\caption{Ablation study on different backbones, namely SAN, S-MRL, and UpDn, as listed here. As listed, our proposed method is improving the results when used with all of the backbones here, and also improves as we use the fusion and causal graph that is proposed. The fusion function is with $\alpha=1.4$ as the free parameter and based on empirical study. }
\label{tab:ablation_backbones}
\centering

\scalebox{1.0}{
\begin{tabular}{|l|rrrr|}
\hline
             & \multicolumn{1}{l}{All}    & \multicolumn{1}{l}{Y/N} & \multicolumn{1}{l}{Num.} & \multicolumn{1}{l|}{Other} \\ \hline
SAN \citep{yang2016stacked} ~~~~~~~~~         & \multicolumn{1}{r}{32.77} & 38.12                   & 12.38                    & 35.56                     \\ \hline
SAN+EA      & 46.25                      & 62.13                   & 37.58                    & 40.31                     \\ \hline
+PW-VQA (EA) & 57.06                      & 86.4                    & 57.73                    & 41.57                     \\ \hline
\end{tabular}
}
\hfill
\scalebox{1.0}{
\begin{tabular}{|l|rrrr|}
\hline
             & \multicolumn{1}{l}{All} & \multicolumn{1}{l}{Y/N} & \multicolumn{1}{l}{Num.} & \multicolumn{1}{l|}{Other} \\ \hline
UpDn \citep{anderson2018bottom} ~       & 37.55                   & 42.11                   & 12.88                    & 41.93                     \\ \hline
UpDn (EA)   & 47.02                   & 65.89                   & 18.11                    & 45.06                     \\ \hline
+PW-VQA (EA) & 58.64                   & 89.51                   & 57.15                    & 45.68                     \\ \hline
\end{tabular}
}
\hfill
\scalebox{1.0}{
\begin{tabular}{|l|rrrr|}
\hline
             & \multicolumn{1}{l}{All} & \multicolumn{1}{l}{Y/N} & \multicolumn{1}{l}{Num.} & \multicolumn{1}{l|}{Other} \\ \hline
S-MRL \cite{cadene2019rubi}        & 36.59                   & 40.71                   & 13.17                    & 40.85                     \\ \hline
S-MRL (EA)   & 49.65                   & 72.48                   & 24.42                    & 44.6                      \\ \hline
+PW-VQA (EA) & 59.54                   & 87.95                   & 59.05                    & 45.83                     \\ \hline
\end{tabular}
}
\vspace{-4mm}
\end{table*}

\subsection{Effect of $\alpha$ variations on the performance}
Ablation study of PW-VQA $\alpha$ values on the final performance result for $\alpha$ values ranging from 1 to 2 are listed in Tab. \ref{tab:ablation_alpha_effect}. As listed and seen in the table, the $\alpha=1.4$ works well for most of the backbones.
\begin{table*}[t]
\centering
\caption{Ablation of PW-VQA $\alpha$ values on the final result for values ranging from 1 to 2. As shown the value of $\alpha=1.4$ works well for most of the backbones.}
\label{tab:ablation_alpha_effect}
\centering

\scalebox{0.71}{
\begin{tabular}{r|rrrr}
\hline
\multicolumn{1}{l}{}    & \multicolumn{1}{l}{All} & \multicolumn{1}{l}{Y / N} & \multicolumn{1}{l}{Num} & \multicolumn{1}{l}{Other} \\ \hline
\multicolumn{1}{l}{SAN} & 33.18                   & 38.57                     & 12.25                   & 36.1                      \\ \hline
$\alpha = $1                       & 56.23                   & 86.2                      & 57.72                   & 40.3                      \\ \hline
$\alpha = $1.1                     & 56.27                   & 87.45                     & 58.5                    & 39.54                     \\ \hline
$\alpha = $1.2                     & 56.96                   & 86.84                     & 58.07                   & 41.57                     \\ \hline
$\alpha = $1.3                     & 52.9                    & 76.87                     & 57.67                   & 39.95                     \\ \hline
$\alpha = $1.4                     & 57.06                   & 86.4                      & 57.73                   & 41.57                     \\ \hline
$\alpha = $1.5                     & 42.75                   & 51.24                     & 52.24                   & 39.38                     \\ \hline
$\alpha = $1.6                     & 55.1                    & 85.64                     & 58.31                   & 38.39                     \\ \hline
$\alpha = $1.7                     & 56.2                    & 86.41                     & 58.21                   & 41.15                     \\ \hline
$\alpha = $1.8                     & 53.85                   & 87.36                     & 54.65                   & 39.3                      \\ \hline
$\alpha = $1.9                     & 43.41                   & 73.47                     & 57.52                   & 34.49                     \\ \hline
$\alpha = $2                       & 53.39                   & 86.22                     & 52.64                   & 39.26                     \\ \hline
\end{tabular}
}
\hfill
\scalebox{0.71}{
\begin{tabular}{r|rrrr}
\hline
\multicolumn{1}{l}{}     & \multicolumn{1}{l}{All} & \multicolumn{1}{l}{Y / N} & \multicolumn{1}{l}{Num} & \multicolumn{1}{l}{Other} \\ \hline
\multicolumn{1}{l}{UpDn} & 37.69                   & 43.17                     & 12.53                   & 41.72                     \\ \hline
$\alpha = $1                        & 57.75                   & 89.09                     & 53.25                   & 45.05                     \\ \hline
$\alpha = $1.1                      & 58.45                   & 89.8                      & 55.5                    & 45.86                     \\ \hline
$\alpha = $1.2                      & 57.55                   & 89.22                     & 57.24                   & 43.08                     \\ \hline
$\alpha = $1.3                      & 57.64                   & 89.24                     & 54.1                    & 45.67                     \\ \hline
$\alpha = $1.4                      & 58.64                   & 89.51                     & 57.15                   & 45.68                     \\ \hline
$\alpha = $1.5                      & 59.13                   & 89.34                     & 57.71                   & 45.38                     \\ \hline
$\alpha = $1.6                      & 58.59                   & 88.18                     & 58.08                   & 45.2                      \\ \hline
$\alpha = $1.7                      & 56.96                   & 89.07                     & 45.16                   & 44.78                     \\ \hline
$\alpha = $1.8                      & 57.09                   & 89.34                     & 54.69                   & 44.83                     \\ \hline
$\alpha = $1.9                      & 58.91                   & 88.51                     & 59.66                   & 44.14                     \\ \hline
$\alpha = $2                        & 58.67                   & 88.16                     & 59.84                   & 43.95                     \\ \hline
\end{tabular}
}
\hfill
\scalebox{0.71}{
\begin{tabular}{r|rrrr}
\hline
\multicolumn{1}{l}{}      & \multicolumn{1}{l}{All} & \multicolumn{1}{l}{Y / N} & \multicolumn{1}{l}{Num} & \multicolumn{1}{l}{Other} \\ \hline
\multicolumn{1}{l}{S-MRL} & 37.09                   & 41.39                     & 12.46                   & 41.6                      \\ \hline
$\alpha = $1                         & 59.47                   & 88.57                     & 58.55                   & 45.44                     \\ \hline
$\alpha = $1.1                       & 59.49                   & 89.1                      & 58.95                   & 45.45                     \\ \hline
$\alpha = $1.2                       & 59.17                   & 87.76                     & 59.53                   & 45.54                     \\ \hline
$\alpha = $1.3                       & 59.24                   & 87.86                     & 59.04                   & 45.71                     \\ \hline
$\alpha = $1.4                       & 59.54                   & 87.95                     & 59.05                   & 45.83                     \\ \hline
$\alpha = $1.5                       & 59.71                   & 88.18                     & 58.31                   & 45.85                     \\ \hline
$\alpha = $1.6                       & 59.44                   & 88.02                     & 58.83                   & 45.43                     \\ \hline
$\alpha = $1.7                       & 59.42                   & 87.79                     & 59.27                   & 45.24                     \\ \hline
$\alpha = $1.8                       & 59.26                   & 87.5                      & 59.56                   & 44.8                      \\ \hline
$\alpha = $1.9                       & 58.82                   & 87.11                     & 59.29                   & 44.31                     \\ \hline
$\alpha = $2                         & 58.4                    & 86.51                     & 57.55                   & 43.99                     \\ \hline
\end{tabular}
}
\vspace{-4mm}
\end{table*}

\end{document}